\documentclass[paper]{ieice}
\pdfoutput=1
\usepackage{graphicx}
\usepackage[fleqn]{amsmath}
\usepackage[varg]{txfonts}
\usepackage{color}

\field{D}
\vol{100}
\no{6}
\title{Construction of Latent Descriptor Space and Inference Model of Hand--Object Interactions}
\authorlist{
 \authorentry[matsuo@i.ci.ritsumei.ac.jp]{Tadashi Matsuo}{m}{ritsumeikanuniv}\MembershipNumber{0224839}
 \authorentry{Nobutaka Shimada}{m}{ritsumeikanuniv}\MembershipNumber{9713662}
}
\affiliate[ritsumeikanuniv]{The author is with the
\EICdepartment{College of Information Science and Engineering}
\EICorganization{Ritsumeikan University}
\EICaddress{1-1-1 Noji-higashi, Kusatsu, Shiga, JAPAN}
}

\received{2016}{9}{27}
\revised{2017}{2}{1}
\onlinefirst{2017}{3}{13}
\jalcdoi{DOI:10.1587/transinf.2016EDP7410}

\newcommand{\revisedregion}{}

\newcommand{\argmin}{\mathop{\rm argmin}\limits}

\newcommand{\defineequal}{\overset{\mathrm{def}}{=}}

\newcommand{\dia}{\mathop{\rm dia}\limits}

\newcommand{\norm}[1]{\left\| #1 \right\|}

\begin{document}
\maketitle
\begin{summary}
 Appearance-based generic object recognition is
 a challenging problem because
 all possible appearances of objects cannot be registered, especially as
 new objects are produced every day.
 {\em Function of objects}, however, has a comparatively small number of
 prototypes.
 Therefore, function-based classification
 of new objects could be a valuable tool
 for generic object recognition.
 Object functions are closely related to hand--object
 interactions during handling of a functional object;
 i.e., how the hand approaches the object, which parts of the object and
 contact the hand, and the shape of the hand during interaction.
 Hand--object interactions are helpful for modeling object functions.
 However, it is difficult to assign discrete labels to interactions
 because an object shape and grasping hand--postures intrinsically have
 continuous variations.
 To describe these interactions, we propose the
 {\em interaction descriptor space}
 which is acquired from unlabeled appearances of human
 hand--object interactions.
 By using interaction descriptors, we can numerically describe the
 relation between an object's appearance and its possible
 interaction with the hand.
 The model infers the quantitative state of the interaction
 from the object image alone.
 It also identifies the parts of objects designed for hand interactions
 such as grips and handles.
 We demonstrate that the proposed method can unsupervisedly generate
 interaction descriptors that make clusters corresponding to interaction
 types.
 And also we demonstrate that the model can infer possible hand--object
 interactions.
\end{summary}

\section{Introduction}
 Appearance-based generic object recognition is
 a challenging problem because all possible appearance of new objects,
 which are produced every day
 cannot be completely registered.
 In contrast, {\em function of objects} is common
 to many objects regularly handled by humans and has a
 comparatively small number of prototypes.
 Therefore, a function-based classification
 of new objects could be valuable
 for generic object recognition.
 {\revisedregion
 The effectiveness of object functions in generic object recognition
 has been already discussed and indicated
 \cite{99242,bowyer2009object}.
 However, in the papers, each function is manually defined for each
 object category.
 It is desirable that information specifying functions can be extracted
 without manually assigning function labels to many objects.
 }

\begin{figure}[t]
  {\centering
  \includegraphics[width=.35\textwidth]{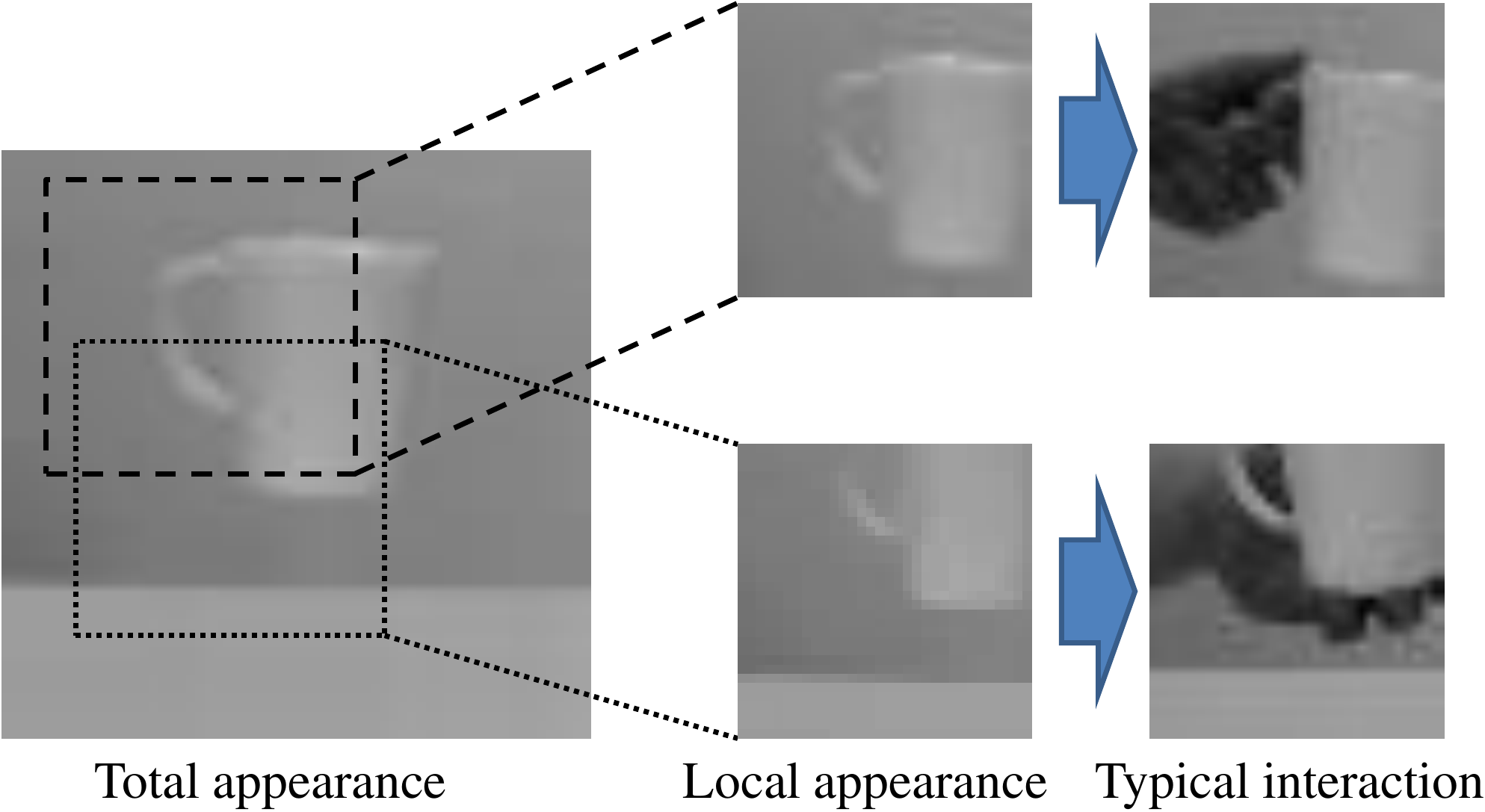}
  \caption{Local appearances and typical hand--object interactions of a cup}
  \label{fig:local_appearances_and_typical_interactions}
  }
 \end{figure}

 The object function is closely related to the
 interactions between the functional object and human hand.
 Specifically, it embodies the approach of the hand to the object,
 the parts of the object contacted by the hand, and the hand shape
 activated by the interaction
 \cite{bub2006:doi:10.1080/02687030600741667}.
 The interaction types specified by such factors have been precisely
 analyzed in the
 literature \cite{napier1956prehensile}.
 Hand--object interactions are therefore promising
 for function-based classification in image-based recognition.
 Some parts of objects, such as grips, bottoms, and brims,
 are handled in typical ways
 (Fig.~\ref{fig:local_appearances_and_typical_interactions}).
 Such interactions with specific parts are
 called {\em perceived affordance} \cite{Norman:1999:ACD:301153.301168}.

 Assuming that a hand--object interaction can be represented by a
 descriptor,
 the descriptor can be considered as a latent attribute of the object
 itself.
 Such descriptors are available for training samples but not for the
 test samples.
 In the context of machine learning,
 training with {\em hidden information} (such as latent attributes)
 can improve the
 classification accuracy
 \cite{6976761,6976958,6226404,Vapnik2009544}.
 The hidden information contains additional records of each training sample:
 for example, age, gender, or race in facial recognition algorithms.
 The classifier is trained to recognize facial image patterns by
 calculating the
 similarity metric of the hidden information such as age, gender, or race,
 which provides the error costs.
 Although the hidden information is not available for test samples,
 considering the hidden information on training brings
 the classification boundaries
 with no over-fitting and good inference performance.
 A similar framework is potentially applicable to recognition based on
 hand--object interactions.

 Alessandro Pieropan et al. have proposed an method estimating an
 object function from a sequence of interactions\cite{6630736}.
 They focus on defining a function by a sequence of descriptions of
 predefined actions (comparatively large motion), not including
 shapes of hands and objects, which are important for interaction
 between a hand and a tool.
 Dan Song et al. have proposed an method estimating a human intention
 from an image sequence of an interaction\cite{6630785}.
 In the method, relationship between intention and appearances is
 learned supervisedly.
 The interaction type is discretely defined and required to be given
 manually before modeling for each action samples.
 Since an object shape and grasping hand--postures intrinsically have
 continuous variations,
 descriptions of interactions should reflect such variation
 continuously.

 We propose a system that can embed a hand--object interaction as a
 ``interaction descriptor'' vector in a small dimensional space.
 The interaction descriptor represents hand--object interactions
 continuously
 in contrast to discrete label sets, which only discriminate several
 predefined objects (``cup'', ``pen'', ...) or function classes (for
 ``drink'', ``write'', ...).
 The proposed method can achieve the embedding in unsupervised way.
 Assuming that object function is closely related to hand-object
 interactions, the interaction descriptor is helpful for modeling object
 functions.

 For a numerical representation of hand--object interactions,
 we introduce the {\em interaction descriptor space}.
 This space is unsupervisedly constructed by a
 convolutional autoencoder (CAE) \cite{masci2011scae},
 an unsupervised feature extraction method.
 When training the model, we introduce a sparseness term in the
 evaluation
 function that clusters similar interactions in the descriptor
 space.
 The training is based on the appearances of hand--object interactions
 of
 typical
 functional objects such as scissors, cutters, pens, and cups.
 The latent attributes in the training are the {\em interaction images},
 comprising the
 appearance itself and its segmentation images of the
 hand and object.
 The descriptor space can quantitatively discriminate among an infinite
 number of functional
 object types.

 Employing the convolutional neural network (CNN) \cite{726791},
 we then
 model the relation between an object's appearance and its corresponding
 hand--object interaction state in the interaction descriptor space.
 In this way, the model can infer the interaction state from the object
 image alone.

 We demonstrate that the descriptor space and the proposed framework
 successfully encode the hand--object interaction state from a single
 object image.

  \section{Interaction descriptor space}
 The object function is closely related to the type of hand--object
 interaction during handling of a functional object, such as
 grasping the object, picking it up, the direction of approach of the
 hand, and other characteristic motions.
 Therefore,  these hand--object interactions are potentially useful for
 describing the object function.
 Since an object shape and grasping hand--postures intrinsically have
 continuous variations,
 interaction descriptors should continuously reflect such variations.
 We represent an interaction descriptor as a vector in a continuous
 vector space, ``interaction descriptor space''.
 We generate an interaction descriptor vector by encoding an appearance
 of a hand--object interaction because the appearance reflects the type
 of the interaction.

 The problem is how to generate the mapping from an appearance to an
 interaction descriptor.
 The mapping should satisfy the following conditions:
 \begin{enumerate}
  \renewcommand{\theenumi}{\Alph{enumi}}
  \item \label{item:extract-essential-info}
        The mapping extracts only the essential information of the
        interaction.
        The detailed shape or texture of the object, which are not
        relevant to the interaction,
        should be ignored.
  \item \label{item:learned-from-unlabeled-appearances}
        The mapping can be learned with a set of unlabeled appearances
        and therefore can be generated
        without manually
        classifying the interactions beforehand.
  \item \label{item:different-interaction-gives-different-descriptor}
        Images corresponding to different interactions are mapped
        to distantly spaced descriptors.
        The difference between two interactions should be reflected in
        the numerical
        distance between their corresponding descriptors.
  \item \label{item:be-tolerant-of-small-shift-scaling}
        Images corresponding to similar interactions are mapped to
        closely spaced descriptors,
        even when the objects differ in size or shape and are slightly
        displaced from each other in the images.
  \item \label{item:based-on-features}
        %
        %
        Certain spatially local features, such as edges and grips,
        are common to multiple interactions and are effective for
        distinguishing among interactions.
        Such useful features should be automatically found from a set of
        appearances.
 \end{enumerate}

 The essential information can be extracted by the autoencoder method
 \cite{BALDI198953,DBLP:journals/corr/MakhzaniF13},
 which employs an encoder and a decoder.
 The encoder converts an input to a code with lower dimensionality,
 and the decoder approximately restores the original input
 from the code.
 Both elements are trained such that the combination restores the input
 as
 correctly as possible for a certain set of vectors.
 Under this constraint, the encoder generates a numerical representation
 of the
 principal components required for input restoration.
 In addition, the encoder and decoder can be trained with unlabeled
 vectors (satisfying condition
 \ref{item:learned-from-unlabeled-appearances}, mentioned above).

 If we can restore an interaction appearance from a descriptor,
 then the descriptor contains information of the interaction.
 The mapping that satisfies conditions \ref{item:extract-essential-info}
 and \ref{item:learned-from-unlabeled-appearances} is generated by
 autoencoder method.

 To satisfy condition
 \ref{item:different-interaction-gives-different-descriptor},
 we concentrate the descriptors corresponding to certain types of
 interaction appearances and
 isolate them from descriptors corresponding to other types of
 interactions.
 If a label specifying an interaction type can be assigned to each
 appearance,
 we can further constrain the mapping so that the
 descriptors of two different interaction types are distantly spaced.
 However, to satisfy \ref{item:learned-from-unlabeled-appearances},
 the important components that specify an interaction must be found from
 unlabeled appearances.
 In previous studies, the important components among a set of unlabeled
 vectors have been found by
 {\em sparse coding} methods
 \cite{5539964,Mairal:2009:ODL:1553374.1553463,NIPS2007_3313,Donoho04032003,5206757}.
 However, these methods require additional inequality or equality
 constraint.

 \begin{figure*}[t]
  {\centering
  \begin{tabular}{c}
   \includegraphics[width=.64\textwidth]{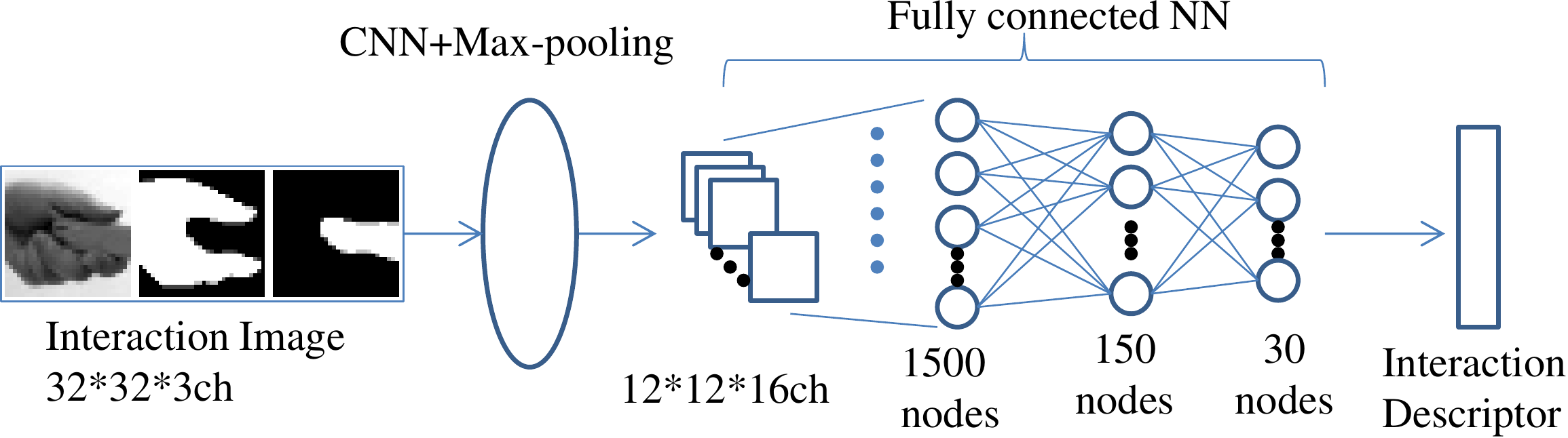}
   \\
   (a) encoder
   \\
   \includegraphics[width=.576\textwidth]{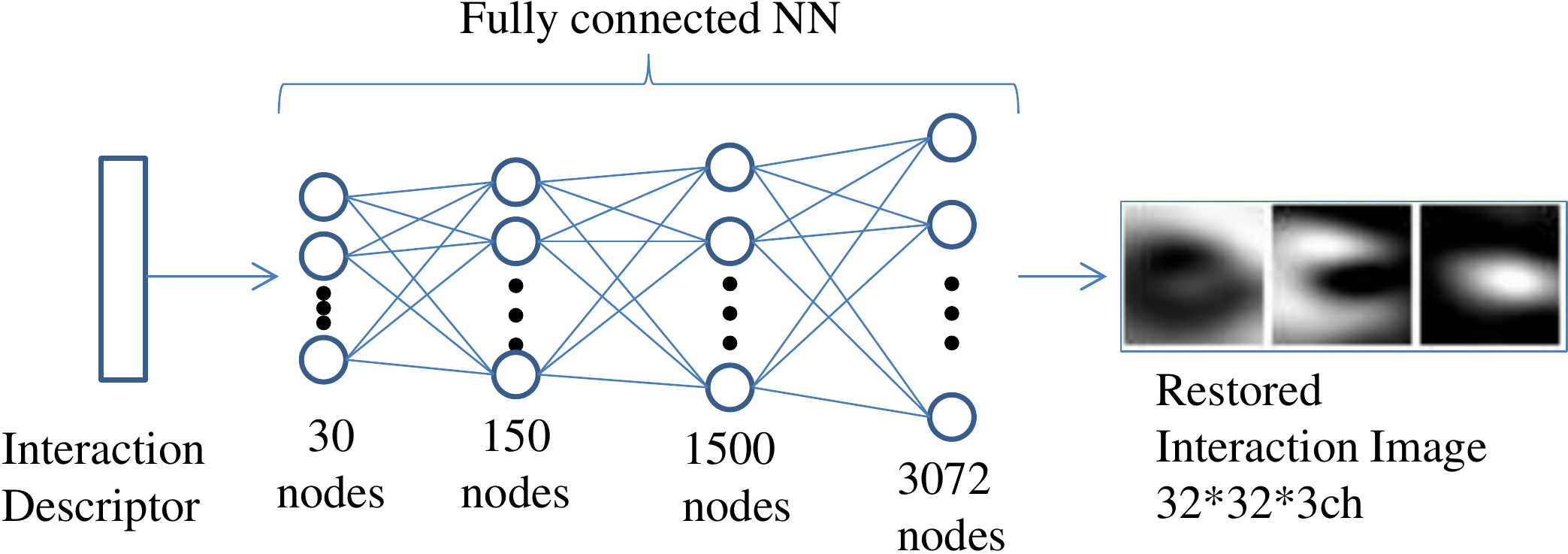}
   \\
   (b) decoder
  \end{tabular}
  \caption{Network structure of the encoder and decoder}
  \label{fig:auto_encoder_network}
  }
  \begin{minipage}[b]{.45\textwidth}
  \begin{center}
   \includegraphics[width=.6\textwidth]{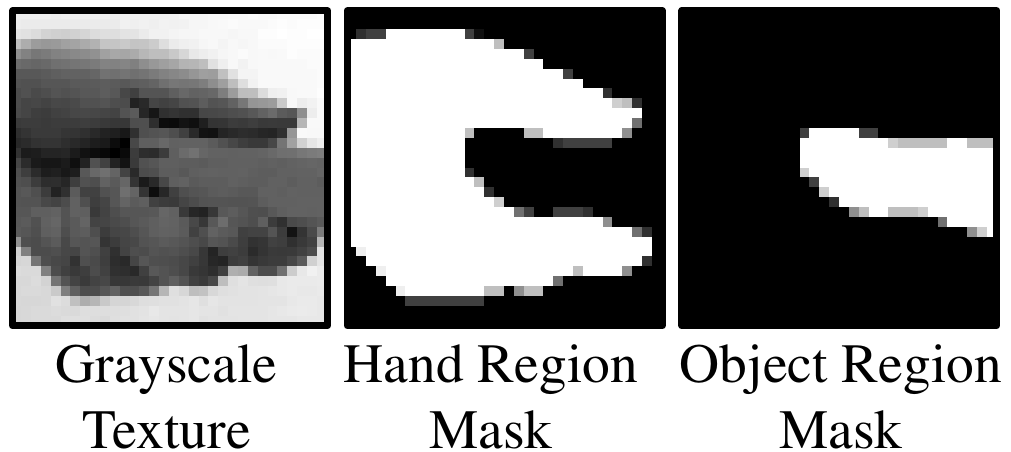}
   \caption{Components of an interaction image}
   \label{fig:interaction_image}
  \end{center}
  \end{minipage}
  \begin{minipage}[b]{.45\textwidth}
  \begin{center}
   \includegraphics[width=.6\textwidth]{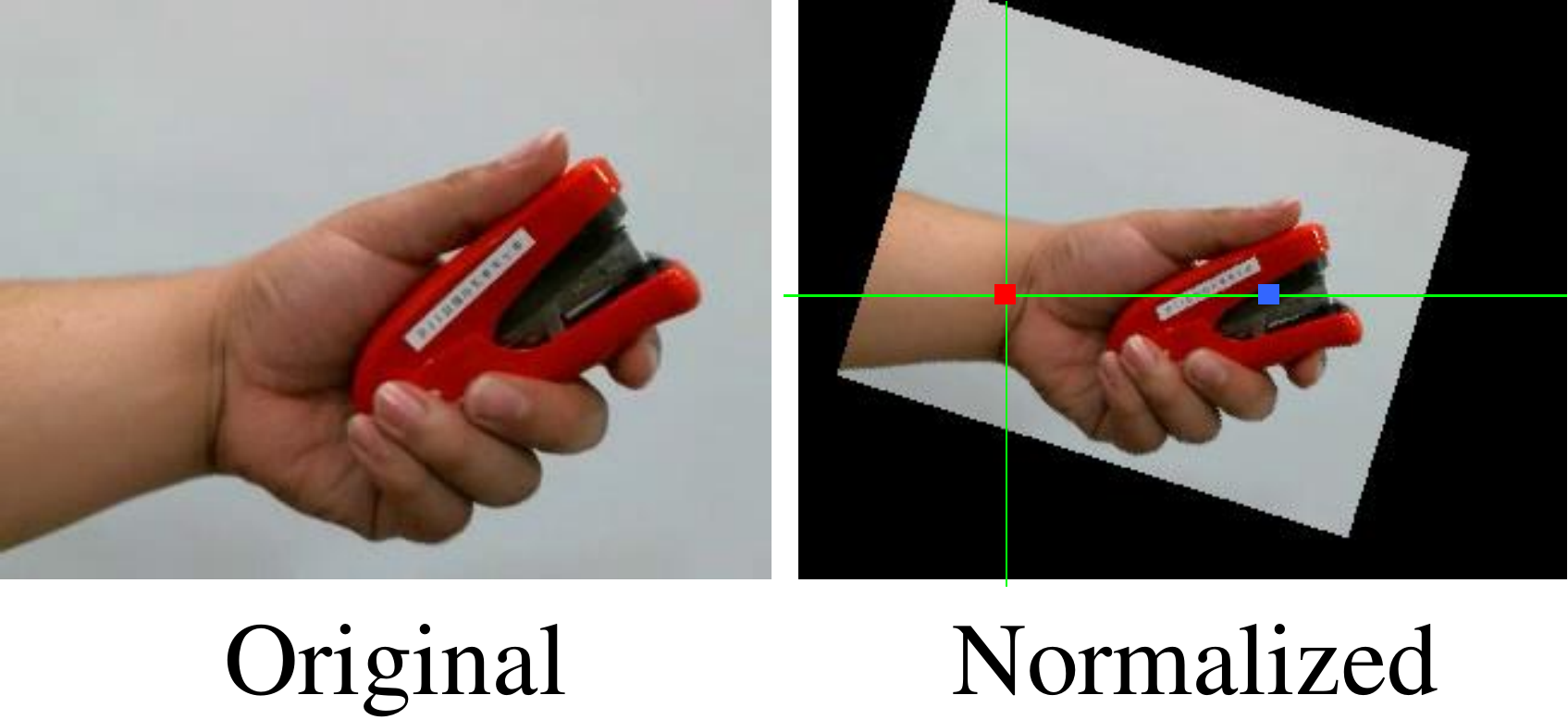}
   \caption{Normalization of a training image}
   \label{fig:normalization_with_wrist_object_coordinate_system}
  \end{center}
  \end{minipage}
 \end{figure*}

 To resolve this problem, we introduce a sparseness constraint to the
 autoencoder.
 Although sparse autoencoders have been previously proposed
 \cite{6618901,DBLP:journals/corr/MakhzaniF13},
 the method in \cite{6618901} requires an inequality constraint
 when training the model.
 The method in \cite{DBLP:journals/corr/MakhzaniF13} requires the
 scheduling of the sparsity level.

 We introduce the sparseness constraint that does not require equality
 or inequality constraint.
 It is applicable to a general CNN-based
 autoencoder that is followed by fully connected neural networks with
 non-linear activation functions.
 As a CNN consists of convolutional filters that uniformly extract
 spatially local features from an image, the extracted local
 features are position-independent (satisfying conditions
 \ref{item:be-tolerant-of-small-shift-scaling} and
 \ref{item:based-on-features}).
 In addition, the CNN filters can be trained by an unsupervised
 learning method
 (satisfying \ref{item:learned-from-unlabeled-appearances}).

 \section{Interaction image}
 Before generating a descriptor from an interaction appearance,
 we need to define the {\em appearance} which contains sufficient
 information
 to distinguish among interactions.

 The appearance is derived from an {\em interaction image}
 (Fig.~\ref{fig:interaction_image}), a 3-channel
 $(32\times 32) \mathrm{pixel}$ normalized image consisting of a total
 appearance, a hand region mask and an object region mask.

 To focus on the hand--object interaction, we approximately normalize the
 positions and directions of the training images for each type of
 hand--object interaction.
 The training images can be automatically normalized in the
 wrist--object coordinate system \cite{morioka2015mprss}, as shown
 in Fig.~\ref{fig:normalization_with_wrist_object_coordinate_system}.
 \begin{figure*}[t]
  {\centering
  \includegraphics[width=.70\textwidth]{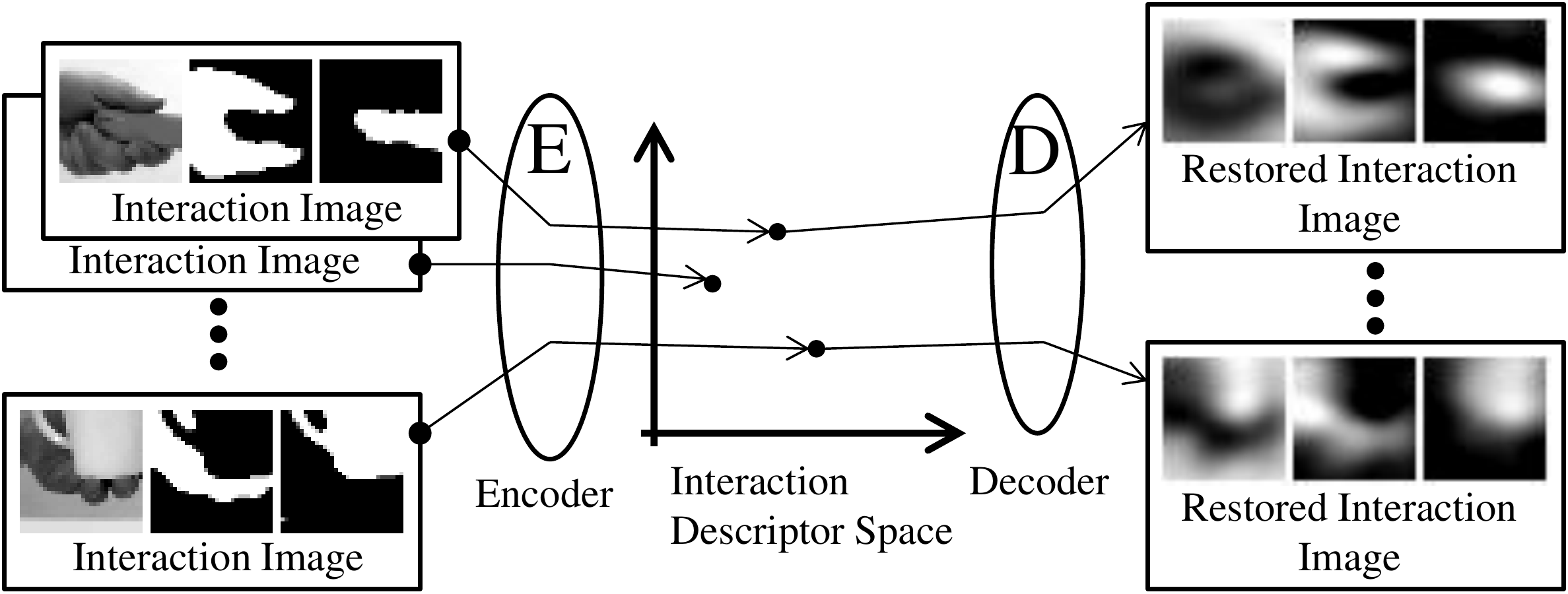}
  \caption{Encoder and decoder outputs}
  \label{fig:encoder_and_decoder}
  }
 \end{figure*}

 \section{Autoencoder for generating descriptors}
 We simultaneously constructed the interaction descriptor space and the
 mapping
 using a sparse convolutional autoencoder (CAE).

 \subsection{Network structure}
 To automatically extract the local features that effectively identify
 an
 interaction, we place the CNN as the first layer of the encoder.
 For learning the nonlinear relation between the local features and
 interactions, the first layer is followed by a three-layer fully
 connected neural network in which each layer precedes a nonlinear
 activation function
 (Fig.~\ref{fig:auto_encoder_network}(a)).
 Similarly, the decoder is a fully connected neural network with a
 nonlinear
 activation function for representing the nonlinear relation
 (Fig.~\ref{fig:auto_encoder_network}(b)).

 \subsection{Cost function}
 Generally, an autoencoder is trained such that the encoder--decoder
 combination
 approximately restores an input in a
 certain input set.
 It is formulated as
 \begin{equation}
  \label{eq:1}
  \argmin_{D, E}
  \sum_{I \in S}\norm{I - D\left(E\left(I\right)\right)}^{2}_{2},
 \end{equation}
 where $I$, $S$, $D( \cdot )$, $E( \cdot )$, and $\norm{\cdot}_{p}$
 denote the input to be reconstructed, a set of inputs, the encoder, the
 decoder, and the
 $\ell^{p}$ norm, respectively.
 In our problem, $I$ and $E\left(I\right)$ denote
 the interaction image and its corresponding descriptor, respectively.

 In the above objective function, the encoder should preserve the
 information of an input among a certain set of inputs.
 Our problem requests that the encoder can extract the essential
 components common to the interaction appearances of similar type.
 According to the basis pursuit concept
 \cite{doi:10.1137/S003614450037906X}
 or sparse coding method,
 this can be achieved by constraining the $\ell^{1}$ norm of the
 encoder's output.
 Simply, the $\ell^{1}$ constraint can be imposed on the
 objective function as follows:
 \begin{equation}
  \beta \sum_{I \in S}\norm{I - D\left(E\left(I\right)\right)}^{2}_{2}
   +
   \lambda
   \sum_{I \in S}
   \norm{E\left(I\right)}_{1}.
 \end{equation}

 However, simply adding the $\ell^{1}$ norm term is ineffective because
 the additional term can be rendered arbitrarily small by scalar
 multiplication
 of the encoder output and the decoder input.
 Basis pursuit avoids this problem by adding a constraint of the
 magnitude of the encoder matrix, as given by Eq.(1) in \cite{6618901}:
 \begin{equation}
  \begin{aligned}
   \argmin_{{\bf D},{\bf z}}&
   \frac{1}{2}\sum_{n=1}^{N}
   \norm{{\bf x}_n - {\bf D}{\bf z}_{n}}_{2}^{2}
   + \beta \norm{{\bf z}_{n}}_{1}\\
   \text{subject to}&
   {\bf D} = [{\bf d}_1, \ldots , {\bf d}_{K}]\\
   &
   \norm{{\bf d}_{k}}_{2}^{2} \leq 1\ \ for \ \ k = 1, \ldots , K,
  \end{aligned}
 \end{equation}
 where ${\bf x}_{n}$ and ${\bf D}$ mean an input vector and an
 {\revisedregion decoder}
 matrix, respectively, and ${\bf z}_{n}$ denotes the code corresponding
 to ${\bf x}_{n}$.
 However,
 such a constraint is not easily imposed on an NN-based encoder.
 Instead, we introduce a constraint term $C_{sparse}$ that
 simultaneously
 limits the $\ell^{1}$ norm and the magnitude of the encoder's output as
 follows:
 \begin{align}
  \label{eq:2}
  C_{err} &=
  \sum_{I \in S_{I}} \norm{I - D\left(E\left(I\right)\right)}^{2}_{2},\\
  C_{sparse} &=
  \sum_{I \in S_{I}}
  \left(
  \frac
  {\norm{E\left(I\right)}_{1}}
  {\norm{E\left(I\right)}_{2}}
  \right)^{2},\\
  C &= \beta C_{err} + \lambda C_{sparse}.
 \end{align}
 The additional term $C_{sparse}$ is the ratio of the $\ell^{1}$ norm
 to the $\ell^{2}$ norm of the descriptor $E\left(I\right)$.
 $C_{sparse}$ is smaller if the descriptor vector $E\left(I\right)$ is
 more sparse \cite{yin2014ratio,DBLP:journals/corr/abs-0811-4706}.
 The autoencoder is trained to minimize the total cost function $C$.

 For a $d$-dimensional vector ${\bf v}$,
 $\left(\norm{\bf v}_{1}/\norm{\bf v}_{2}\right)^{2}$ is
 minimized at 1
 only when the vector ${\bf v}$ has a single non-zero component
 and all other
 components are zero (the sparsest case).
 Conversely, it is maximized when all components of
 ${\bf v}$
 have a common absolute value.
 The lower the $C_{sparse}$, the fewer basis vectors required in the
 weighted sum that approximates the decoder outputs.

 This decoder
 obtains the shapes of a hand and an object and their spatial relation
 from a point on the interaction descriptor space
 (Fig.~\ref{fig:encoder_and_decoder}).

 \section{Inference model}
\begin{figure}[t]
  {\centering
  \includegraphics[width=.45\textwidth]{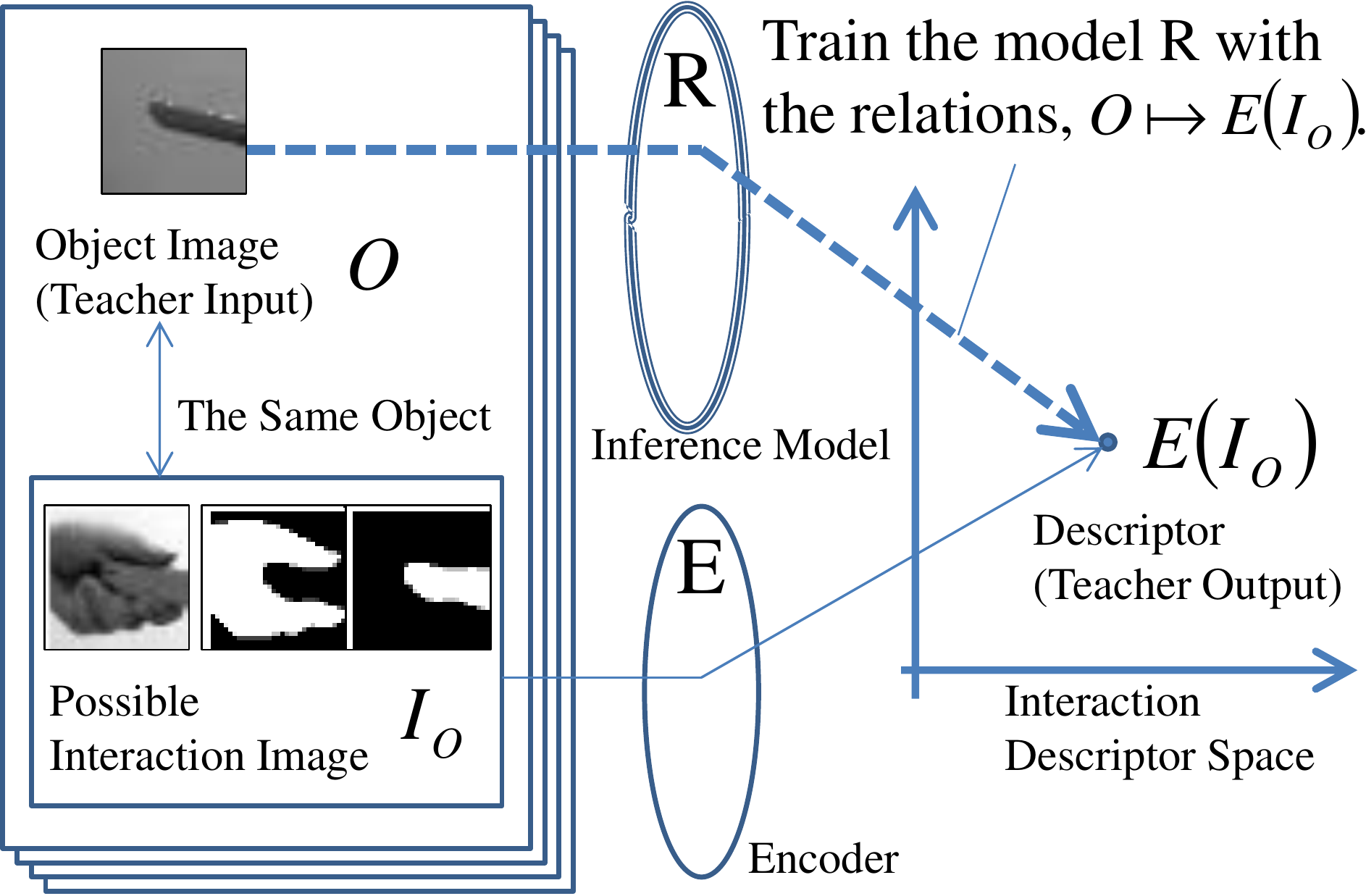}
  \caption{Training of inference model}
  \label{fig:training_of_recollector}
  }
 \end{figure}

 With the numerical interaction descriptor,
 the CNN can learn the relation between an object appearance and a
 possible interaction
 (Fig.~\ref{fig:training_of_recollector}).
 We can then infer an instance of the interaction descriptor
 from the appearance of an object.

 {\revisedregion
 Each interaction descriptor does not represent an absolute direction of
 an interaction in an image because it is based on interaction images
 and they are normalized by the wrist--object coordinate system, as
 shown
 in Fig.~\ref{fig:normalization_with_wrist_object_coordinate_system}.
 An interaction descriptor represents shapes of a hand and an object,
 their relative position and their relative direction.
 So, when pairing an object-only appearance with an interaction
 descriptor for training the inference model,
 any rotated versions of the object-only appearance may be paired
 with the interaction descriptor.
 Although it is possible to train the inference model with all pairs
 of any rotated object-only appearance and an interaction
 descriptor, a convolutional neural network (CNN) cannot effectively
 extract common components from many rotated variations
 \cite{Cheng_2016_CVPR}.
 To learn common shapes of objects with common possible interactions
 by a CNN effectively, it is desirable that directions of objects
 with common interactions are standardized.
 Since an object-only appearance does not have an obvious standard
 direction,
 we have normalized a direction of an object-only appearance so that
 the object has a direction similar to that of an object in the
 paired interaction image.

 Due to the normalization, when inferring an interaction descriptor
 from an object-only appearance, we have to rotate the appearance so
 that its direction matches to that of a similar object used in
 training.
 Since appropriate rotation for an object is unknown generally,
 we have to infer interaction descriptors from all possible
 versions of the rotated appearance.

 However, if objects have some typical poses in images due to
 gravity or other reason, the inference model should be trained with
 pairs of an appearances of an object in such a
 typical pose and a corresponding interaction descriptor.
 The inference model trained in that way can infer an interaction
 descriptor from an object-only appearance itself because it matches to
 one of typical poses.
 }

\begin{figure}[t]
  {\centering
  \includegraphics[width=.45\textwidth]{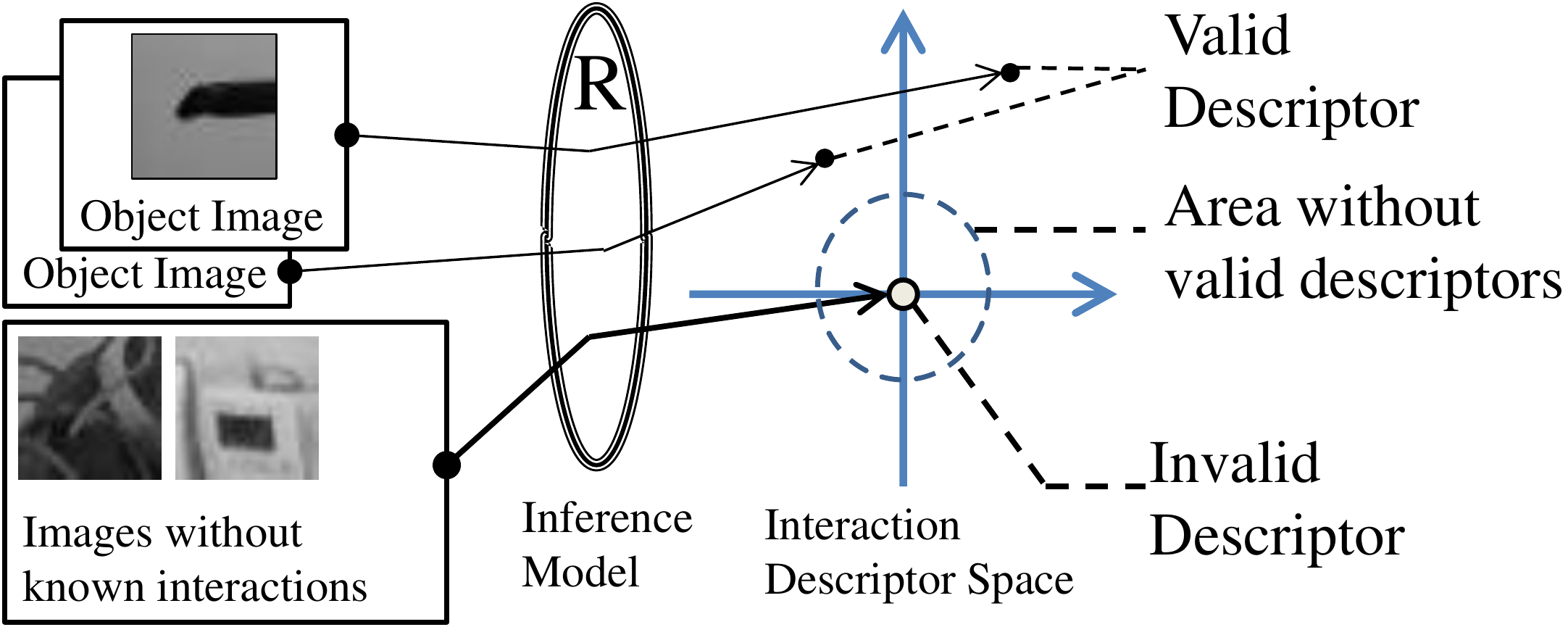}
  \caption{Training with invalid descriptor}
  \label{fig:training_with_invalid_descriptor}
  }
 \end{figure}

 We also introduce an {\em invalid descriptor} that discriminates
 between
 images with and without known interactions.
 For an input without known interactions, the
 model is trained to output the descriptor closest to the invalid
 descriptor.
 The invalid descriptor is defined as a zero vector.
 The model is trained with two types of teacher samples
 (Fig.~\ref{fig:training_with_invalid_descriptor});
 pairs of an image with a known interaction and its descriptor (positive
 samples),
 and pairs of an image without known interactions and an invalid
 descriptor (negative samples).

\begin{figure}
  {\centering
  \includegraphics[width=.45\textwidth]{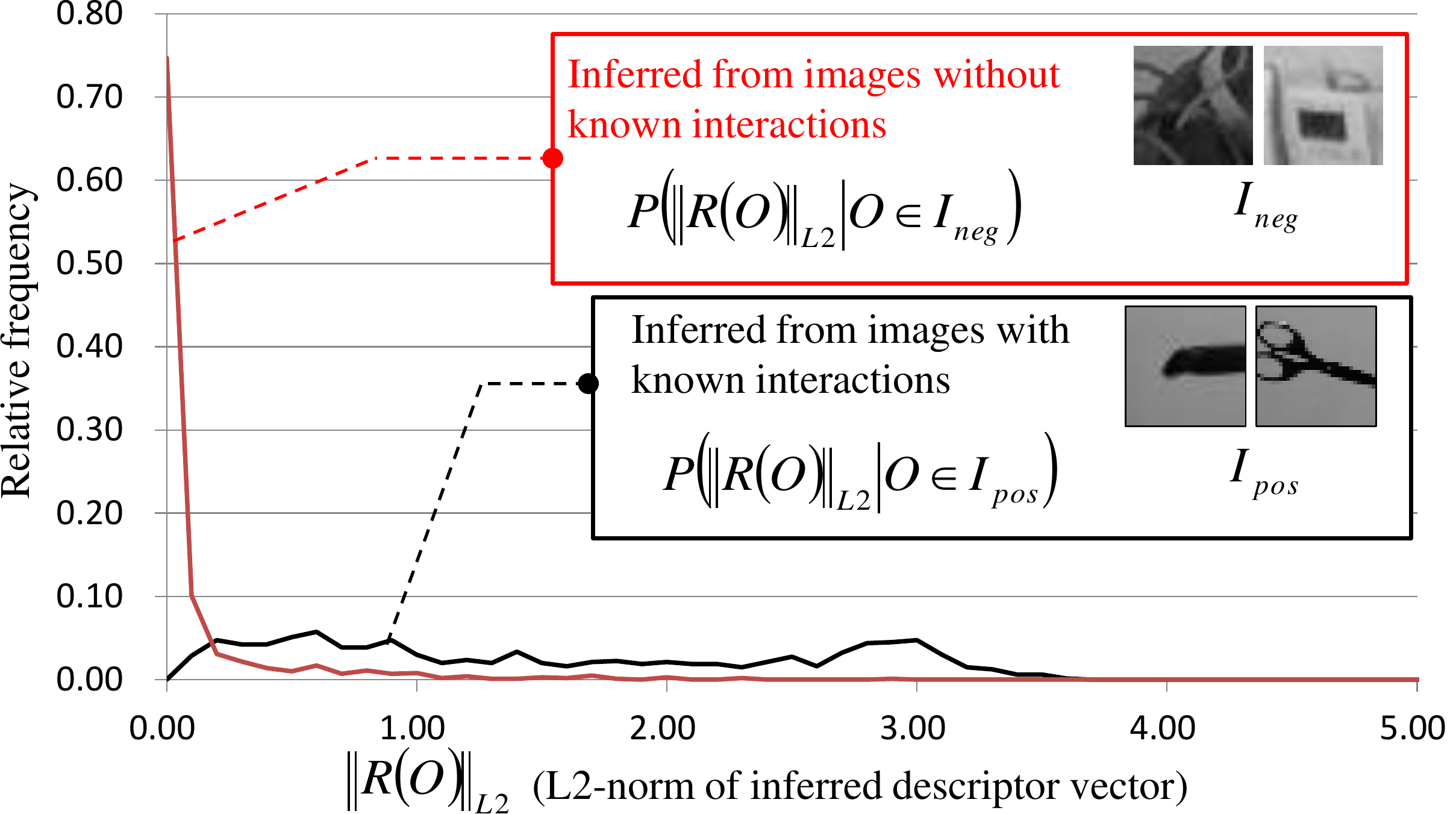}
  \caption{Distribution of $\mathrm{L}_{2}$-norms of inferred descriptors}
  \label{fig:norm_distribution_of_inferred_descriptor}
  }
 \end{figure}

\begin{figure*}[t]
  {\centering
  \includegraphics[width=.64\textwidth]{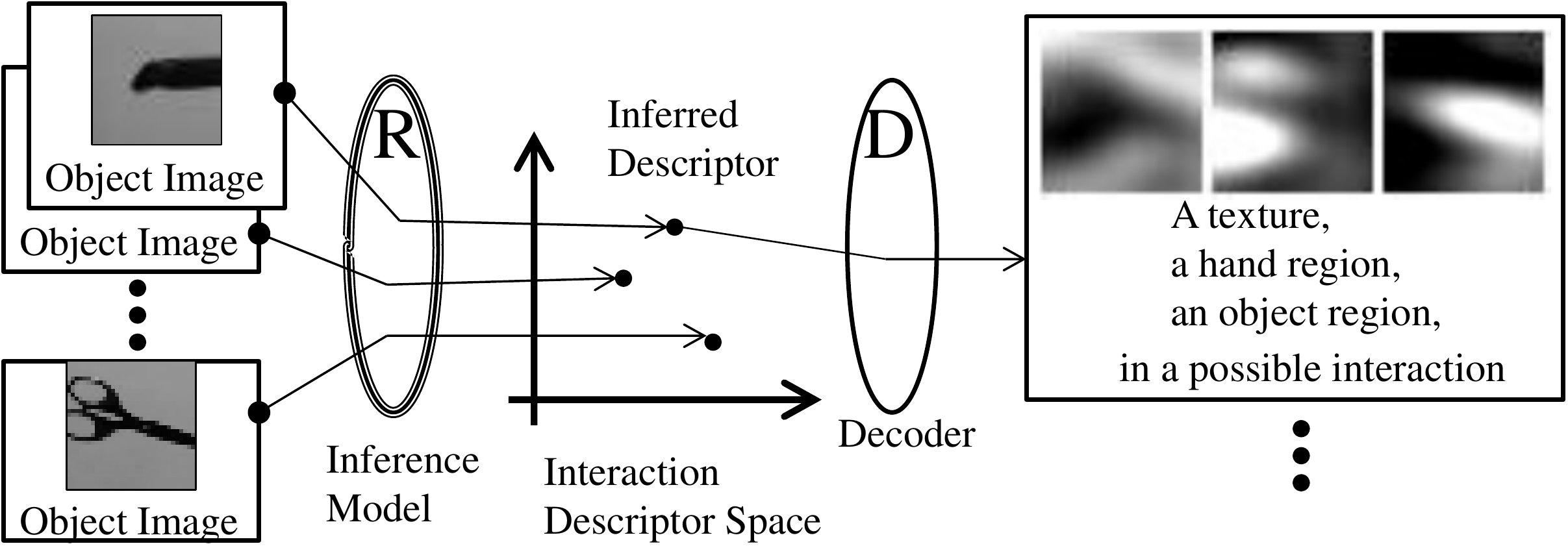}
  \caption{Operation of the inference system}
  \label{fig:recollection_system}
  }
 \end{figure*}

\begin{figure*}
  {\centering
  \includegraphics[width=.64\textwidth]{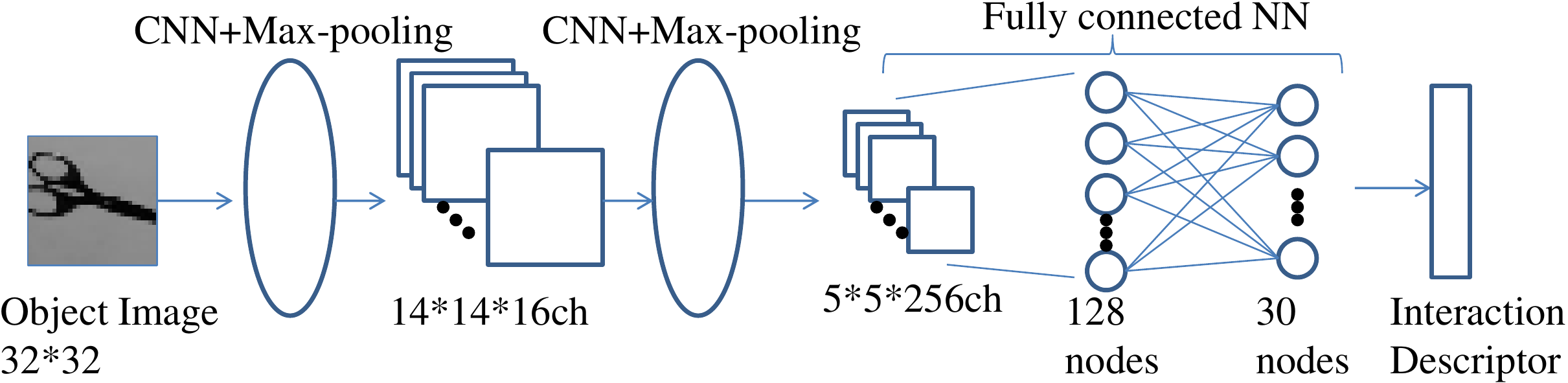}
  \caption{Network structure of the inference model}
  \label{fig:inference_model_network}
  }
 \end{figure*}

 After training the model, we estimate the probability distribution of
 the $\mathrm{L}_{2}$-norms of the inferred descriptors for samples
 of each teacher type.
 These samples differ from the training samples of the
 inference model $R$.
 Figure \ref{fig:norm_distribution_of_inferred_descriptor} shows the
 estimated distributions of
 $P\left(\norm{R(O)}_{\mathrm{L}2}\left|O\in I_{pos}\right.\right)$
 and
 $P\left(\norm{R(O)}_{\mathrm{L}2}\left|O\in I_{neg}\right.\right)$,
 where $R$ denotes the inference model, $O$ denotes an input image,
 and $I_{pos}$ and $I_{neg}$ are the sets of teacher images with and
 without known interactions, respectively.
 As shown in the figure, a high norm of an inferred descriptor indicates
 large likelihood that the input has a known interaction.
 The likelihood $f$ that an input image $O$ has a known
 interaction is given by
 \begin{equation}
  f(O) =
   \frac
   {g(\norm{R(O)}_{\mathrm{L}2})}
   {g(\norm{R(O)}_{\mathrm{L}2})+h(\norm{R(O)}_{\mathrm{L}2})},
 \end{equation}
 where
 \begin{equation}
  \begin{aligned}
   g(d)
   &=
   P\left(d=\norm{R(O')}_{\mathrm{L}2}\left|O'\in I_{pos}\right.\right),
   \\
   h(d)
   &=
   P\left(d=\norm{R(O')}_{\mathrm{L}2}\left|O'\in I_{neg}\right.\right).
  \end{aligned}
 \end{equation}
 By connecting the inference model $R$ and the decoder $D$,
 our system infers a possible interaction image from the
 appearance of an object (Fig.~\ref{fig:recollection_system}).
 The inference model $R$ is the CNN shown in
 Fig.~\ref{fig:inference_model_network}.


 \section{Experiment}
\begin{figure}[t]
  \begin{center}
   \includegraphics[width=.45\textwidth]{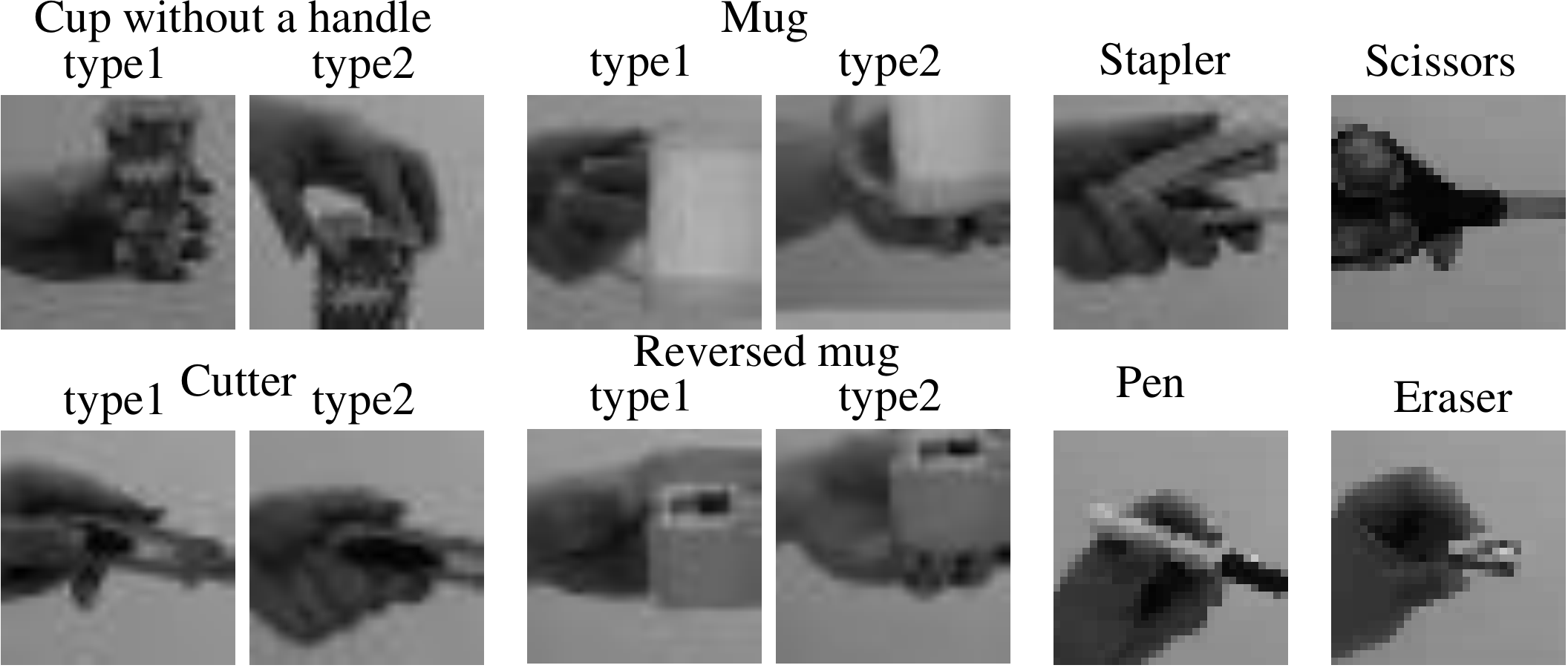}
   \caption{Interaction types in the encoder--decoder training step}
   \label{fig:interaction_type}
  \end{center}
 \end{figure}

 The encoder and decoder were trained with interaction images generated
 from 1,680 scenes showing 12 types of interactions
 (Fig.~\ref{fig:interaction_type}).
 Each interaction image was generated from a
 $(32\times 32) \mathrm{[pixel]}$
 sub-image randomly located in the scene image.
 We generated multiple instances of interaction images
 from each scene by randomly extracting subsquares with sufficient area
 of a hand region.
 The total variation exceeds 500,000.
 Masks in interaction images were generated by skin color extraction and
 background subtraction.
 The encoder and decoder were trained by minimizing $C$ using
 stochastic gradient descent (SGD) \cite{726791}.

 \subsection{Distribution of descriptors}
 To demonstrate the effect of the sparseness cost,
 we define the diameter $\mathcal{D}$ of a set of descriptors as follows:
 \begin{equation}
  \dia \mathcal{D} \defineequal
   \max
   \left\{
    \norm{x-y}
    \mid
    x,y \in \mathcal{D}
   \right\}
 \end{equation}
 We also define $\mathcal{D}_k$ as a set of descriptors of the $k$-th
 interaction type, and denote $\mu_{\text{dia}}$ as the mean of
 $\dia \mathcal{D}_{k}$ for $k$.
 If $\mu_{\text{dia}}$ is small, the
 descriptors corresponding to a similar interaction are closely placed.
 This is a desirable property because similarity of interactions should
 be reflected in closeness of their descriptors.

\begin{figure}
  {\centering
  \includegraphics[width=.45\textwidth]{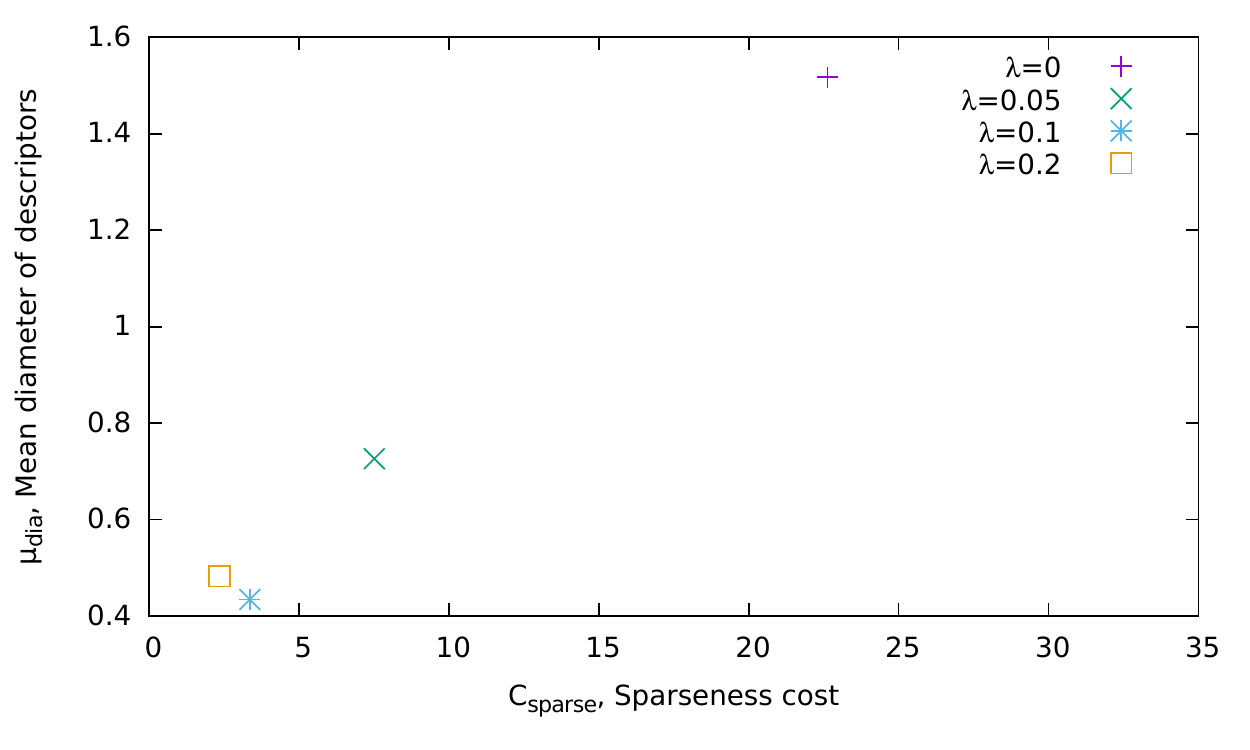}
  \caption{Mean diameter versus sparseness cost of the descriptor groups
  (each symbol denotes a different $\lambda$)}
  \label{fig:cost_and_diameter}
  }
 \end{figure}

 Figure \ref{fig:cost_and_diameter} shows the relation between
 the sparseness cost $C_{sparse}$ and
 the mean diameter $\mu_{\text{dia}}$ of the descriptor sets.
 Each point corresponds to a pair of $C_{sparse}$ and $\mu_{\text{dia}}$
 after the training process for each weight $\lambda$ in the cost
 function (\ref{eq:2}).
 The case $\lambda = 0$ is equivalent to the case without a sparseness
 cost.
 The figure shows that larger $\lambda$ brings smaller $C_{sparse}$ and
 smaller $C_{sparse}$ brings smaller $\mu_{\text{dia}}$.
 Smaller $\mu_{\text{dia}}$ means that
 descriptors corresponding to a similar interaction are more aggregated.
 This result shows that the proposed method can unsupervisedly generate
 interaction descriptors that make aggregates corresponding to
 hand--object interactions.

 To compare distributions of descriptors by autoencoders trained
 with/without the sparseness cost $C_{sparse}$ in (\ref{eq:2}),
 we calculated the purity, an evaluation measure of clustering quality,
 The purity is defined by how many samples in a cluster belong to the
 most frequent class label (correct label given manually) in the
 cluster as follows;
 \begin{equation}
  \text{(PURITY)}
  \defineequal
  \sum_{c: \text{cluster index}}
  \frac{1}{n_{c}}
  \max_{i: \text{interaction type}}{n_{c,i}},
 \end{equation}
 where $n_{c}$ means the number of samples assigned to the $c$-th cluster,
 and $n_{c, i}$ means the number of samples from the $i$-th type
 interaction assigned to the $c$-th cluster.
 If the purity of clusters is close to 1, almost all descriptors of each
 cluster belong to a common interaction type.
 This means that the clusters generated without information of
 interaction types approximately form subdivision of
 interaction types.
 We calculated descriptors from 800 hand--object interaction images
 not used in the training for each autoencoder, and applied mean-shift
 clustering to the descriptors.
 Table \ref{tab:purity} shows the purities for autoencoders trained
 with/without the sparseness cost $C_{sparse}$.
 The purity for the autoencoder trained with the sparseness cost (the
 proposed method) was
 0.99,
 while that for that without sparseness cost was 0.16.
 99 percent of descriptors in a cluster by the proposed method belong to
 a common interaction type.
 This implies that the descriptors for the same interaction type are
 more separately embedded by the autoencoder trained with sparseness
 cost than without sparseness cost.
 It is important that these
 clusters with the similar interaction was unsupervisedly generated
 from only image signals.
 \begin{table}[t]
  \begin{center}
  \caption{Purity of clusters of descriptors}
  \label{tab:purity}
  \begin{tabular}{|c|c|}
   \hline
   Autoencoder& Purity\\
   \hline
   \hline
   Without the sparseness cost& 0.16\\
   \hline
   With the sparseness cost (the proposed method)& 0.99\\
   \hline
  \end{tabular}
  \end{center}
 \end{table}

\begin{figure*}
  {\centering
   \includegraphics[width=.99\textwidth]{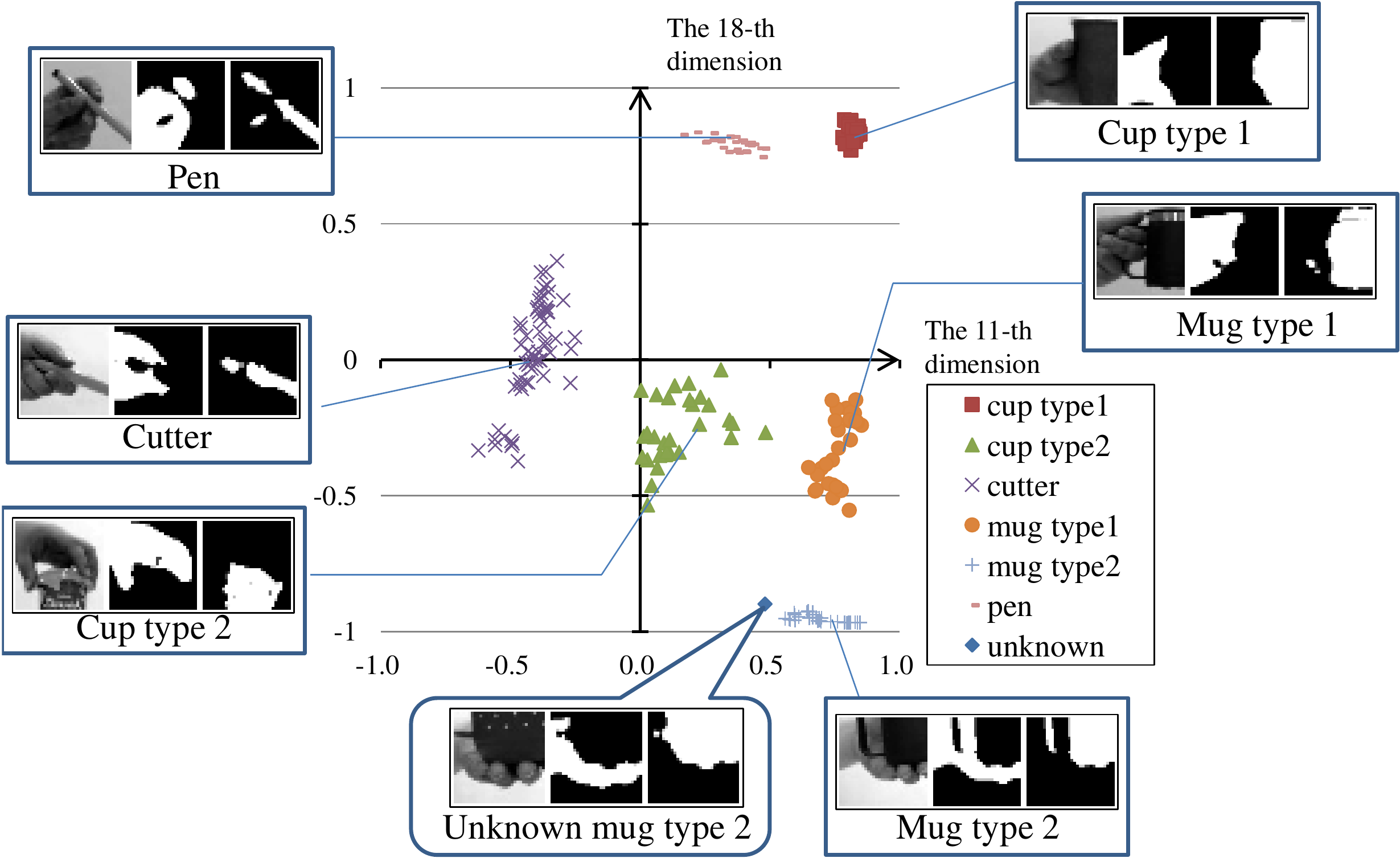}
   \caption{One plane of the interaction descriptor space}
   \label{fig:descriptor_space}
  }
 \end{figure*}

 Figure \ref{fig:descriptor_space} shows the distribution of
 various object images within 11th and 18th dimensions of the
 interaction
 descriptor space.
 ``Mug type 1'' and ``Mug type 2'' are different hand--mug interactions.
 In the former interaction, the hand grips the mug's handle; in the
 latter, the hand holds the mug from the bottom.
 As shown in Fig.~\ref{fig:descriptor_space}, these interactions form
 two separate
 clusters in the descriptor space.
 In addition,
 the interaction image ``Mug type 2'', which was not used in training,
 maps to a descriptor
 near those of the mug--hand interaction images used in training.
 As a group of similar interactions composes a cluster in the
 interaction descriptor space,
 that space well characterizes the types of
 hand--object interactions.

 \subsection{Restoration by the decoder}
\begin{figure}[t]
  {\centering
  \begin{tabular}{cc}
   \includegraphics[width=.215\textwidth]{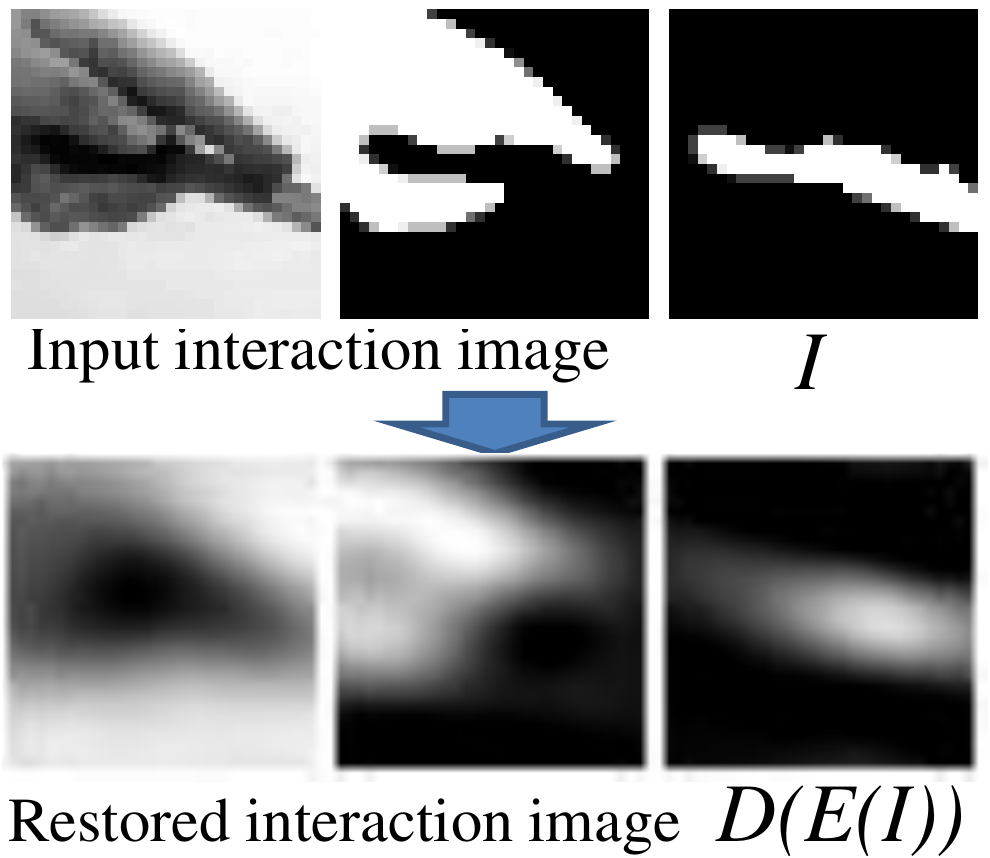}
   &
   \includegraphics[width=.215\textwidth]{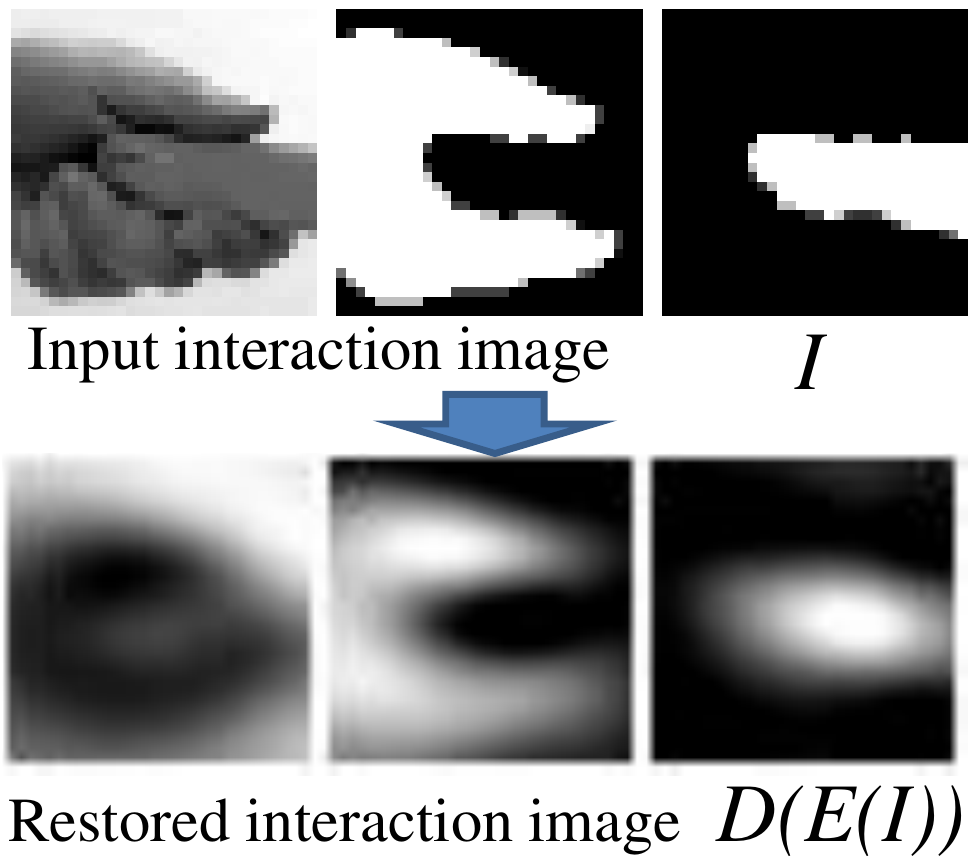}
   \\
   (a) Cutter type1
   &
   (b) Cutter type2
   \\
   \includegraphics[width=.215\textwidth]{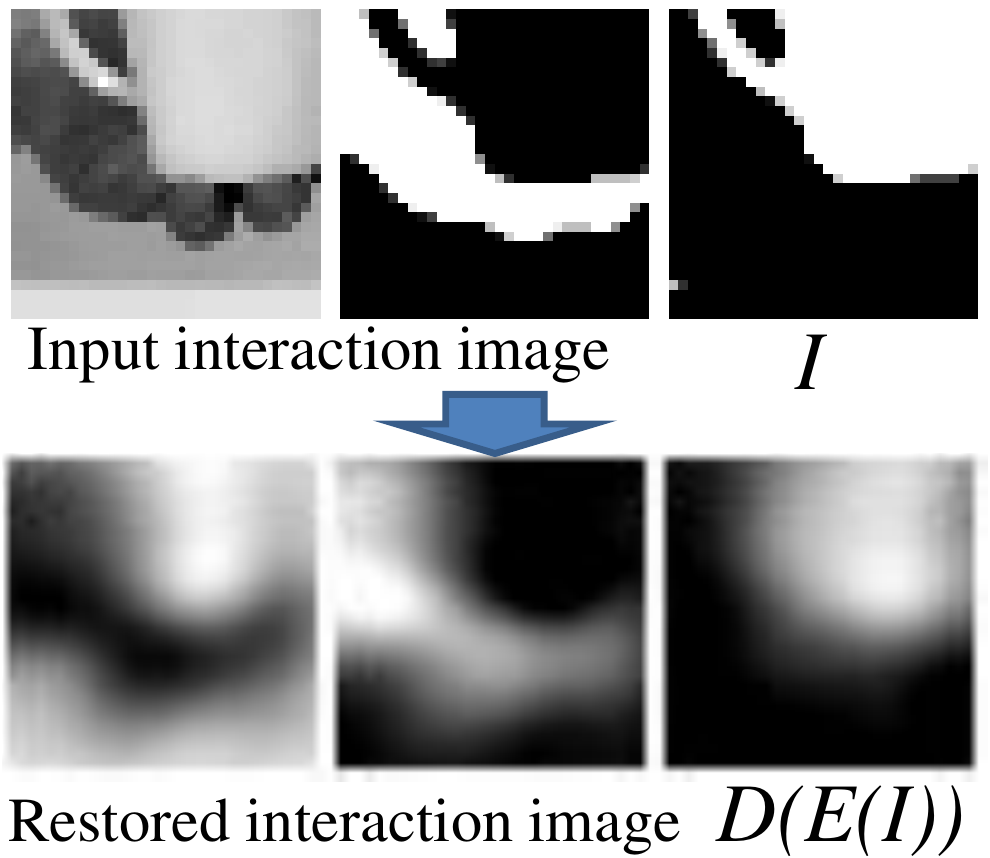}
   &
   \includegraphics[width=.215\textwidth]{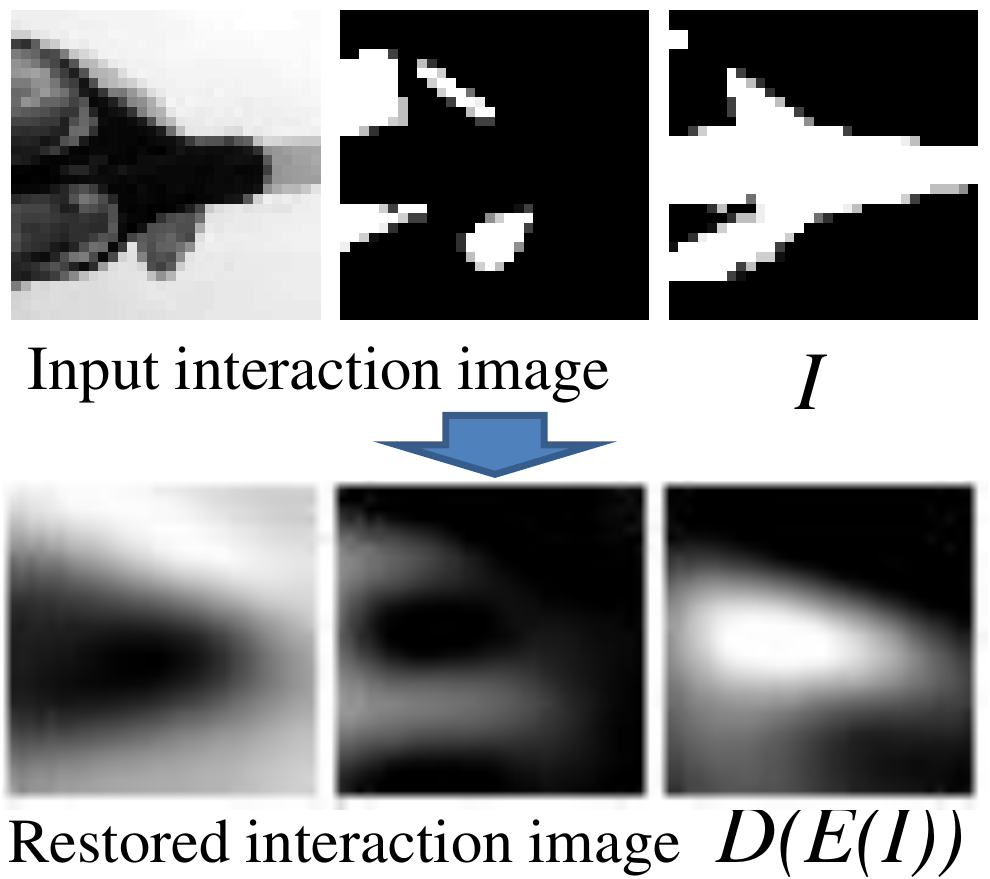}
   \\
   (c) Mug type2
   &
   (d) Scissors
  \end{tabular}
  \caption{Restoration by autoencoder}
  \label{fig:restoration_by_auto_encoder}
  }
 \end{figure}

 Figure \ref{fig:restoration_by_auto_encoder} shows examples of the
 decoder restorations.
 The encoder abstracts the rough shapes and
 positions of image features, ignoring their specific textures.

 \subsection{Inference of an interaction}
\begin{figure}[t]
  {\centering
  \begin{tabular}{cc}
   \includegraphics[width=.215\textwidth]{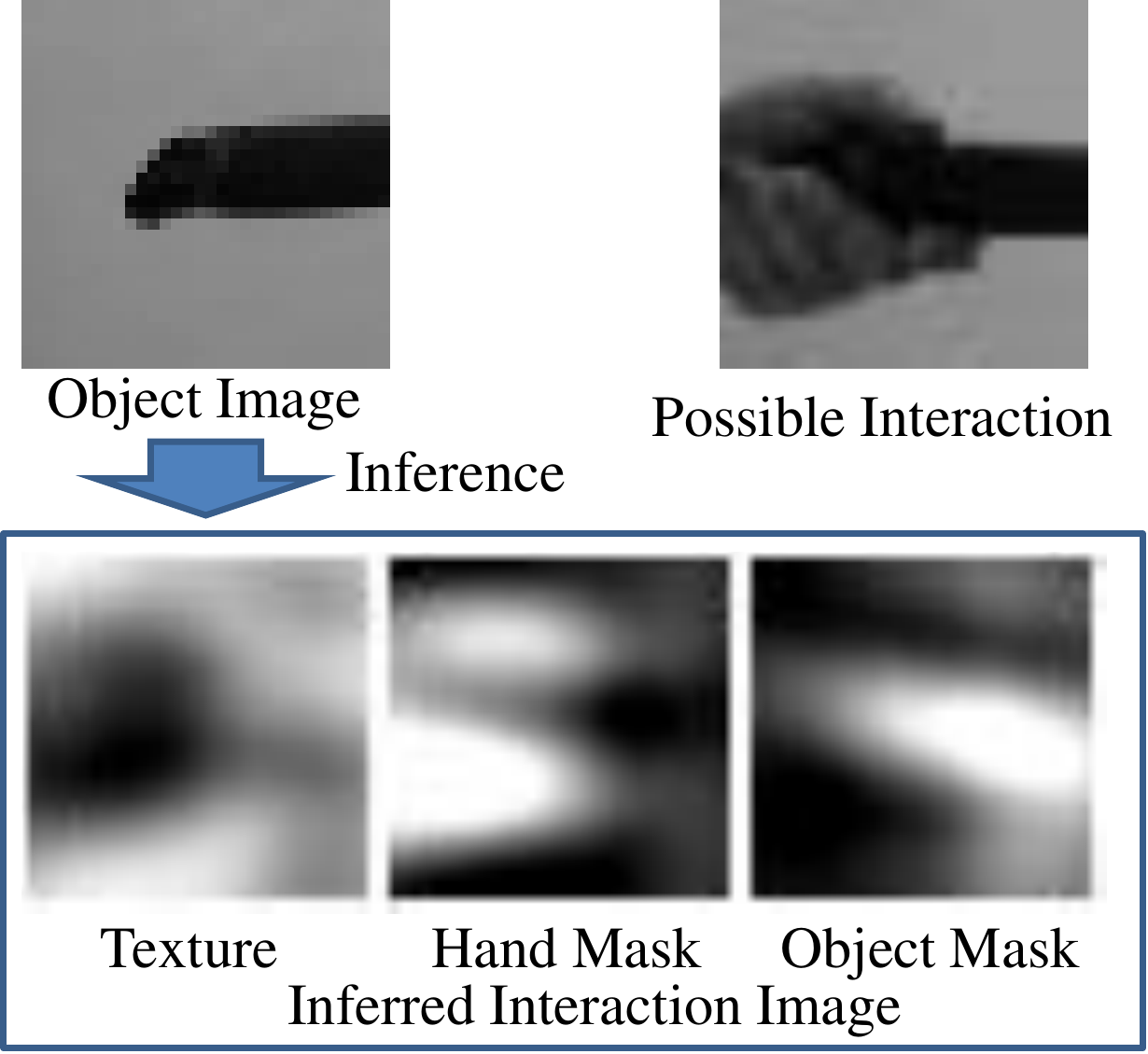}
   &
   \includegraphics[width=.215\textwidth]{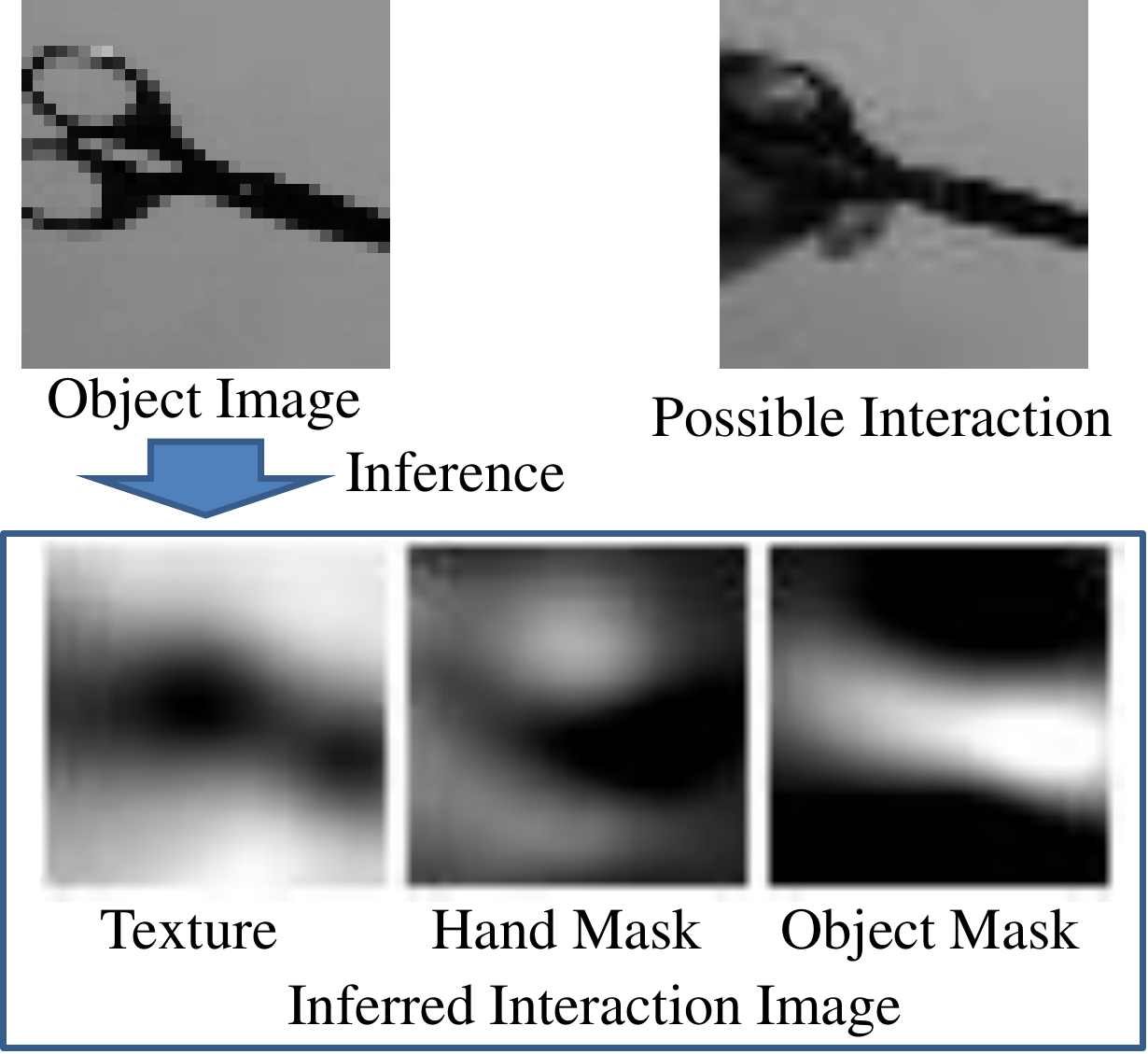}
   \\
   (a) A cutter
   &
   (b) Scissors
   \\
   \includegraphics[width=.215\textwidth]{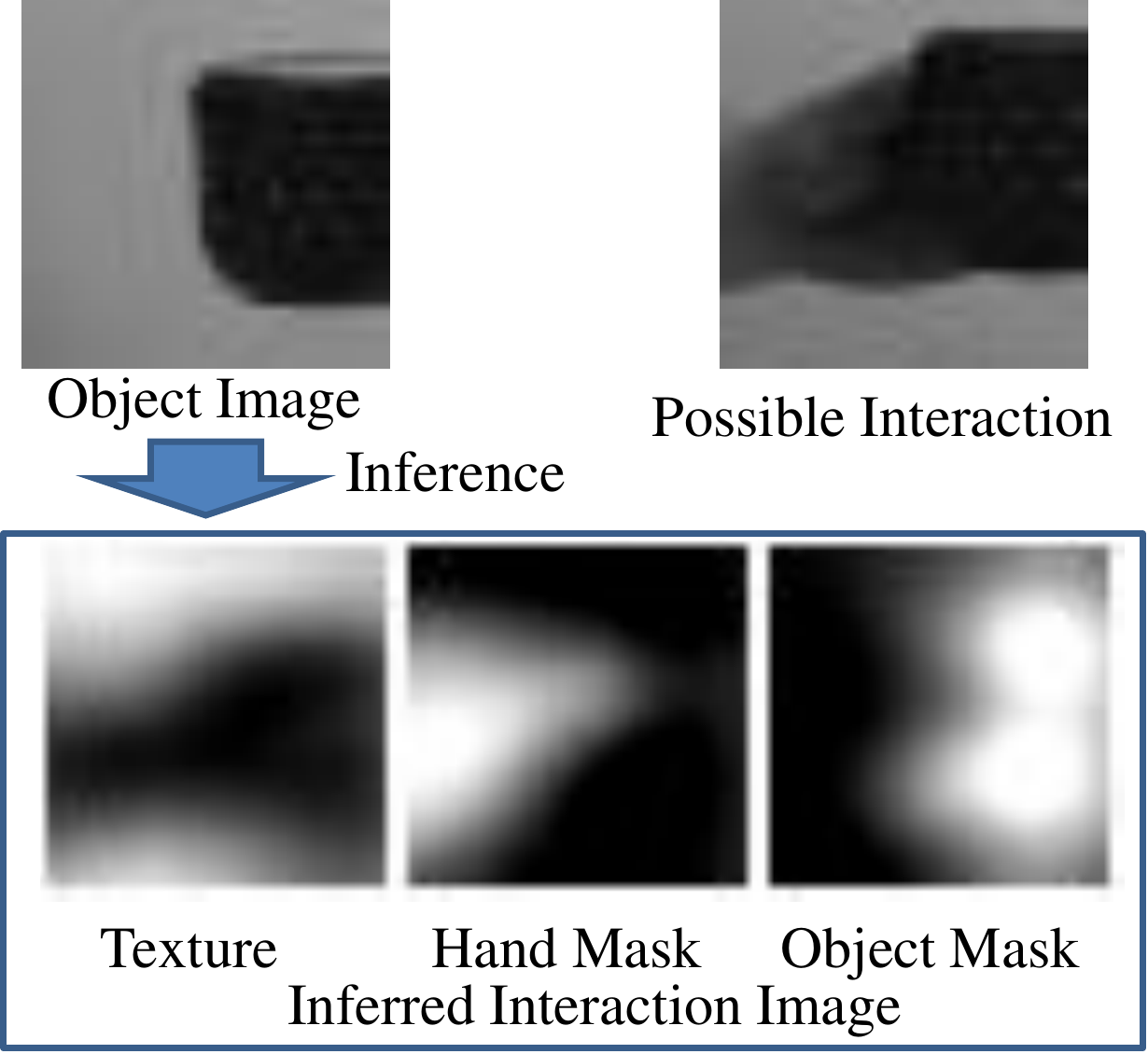}
   &
   \includegraphics[width=.215\textwidth]{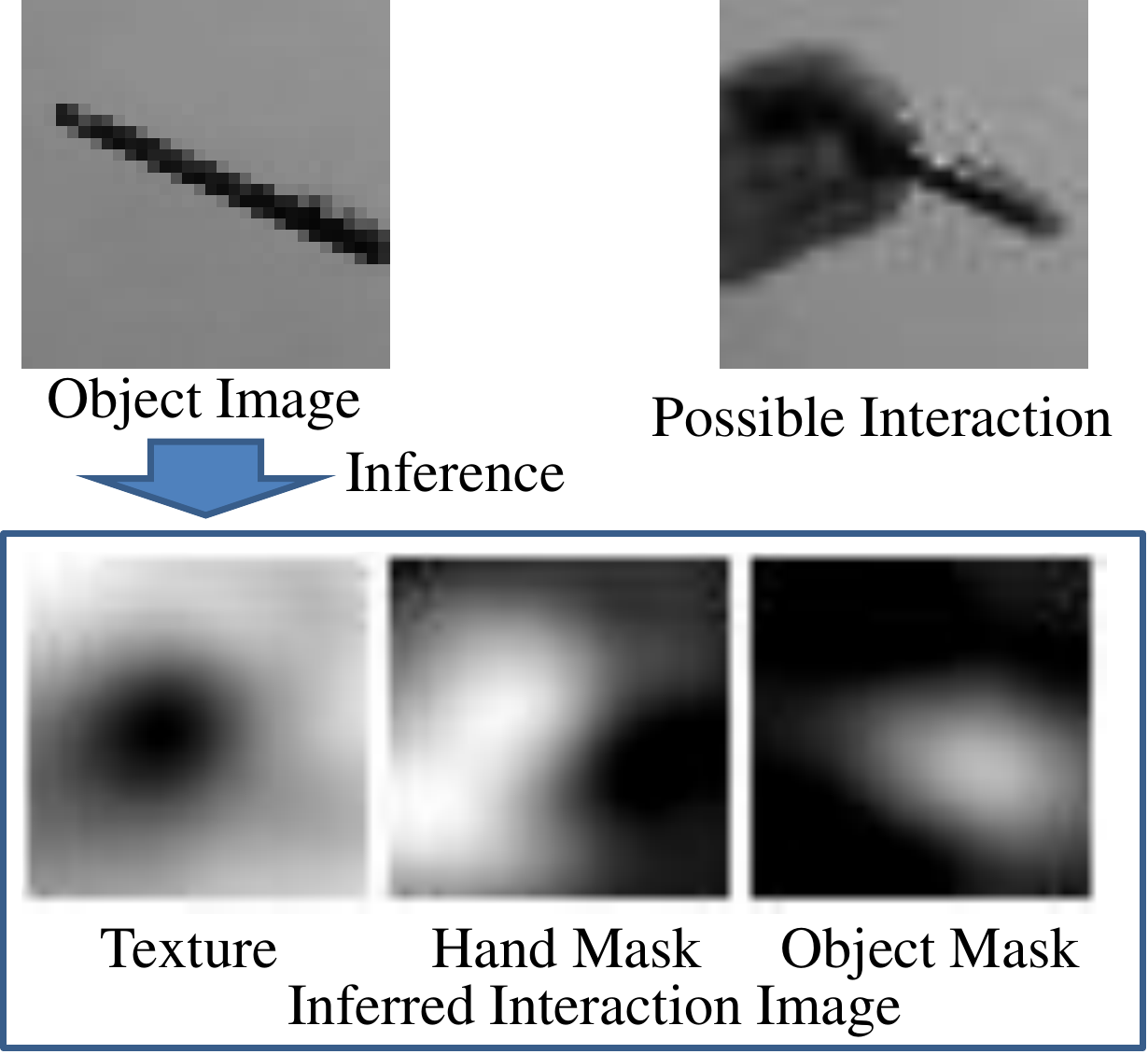}
   \\
   (c) A cup
   &
   (d) A pen
  \end{tabular}
  \caption{Interaction images inferred from unknown objects}
  \label{fig:inference_result}
  }
 \end{figure}

 Figure \ref{fig:inference_result} shows interaction images inferred
 from appearances of objects.
 In these examples, the pairs of masks of the inferred hand
 and object show their positional relations in the possible
 interaction.
 The hand--object interactions were successfully inferred from the
 single
 object images.

\begin{figure}[t]
  {\centering
  \includegraphics[width=.32\textwidth]{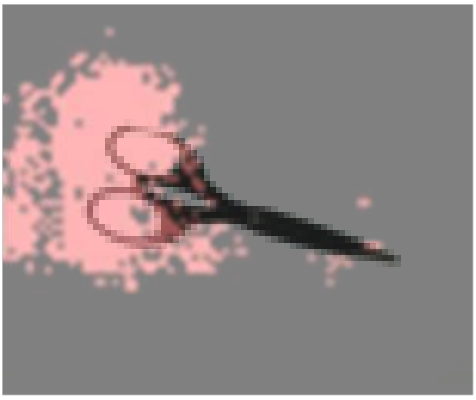}
  \caption{Regions (pink) of possible interaction between human hand and
  scissors}
  \label{fig:map_of_high_likelihood}
  }
 \end{figure}

 Figure \ref{fig:map_of_high_likelihood} shows the region of interaction
 between scissors and a
 human hand.
 The colored regions mark the center of a window in which the likelihood
 $f$ exceeds $0.9$.
 High and low likelihoods are inferred around the grip and edges of the
 scissors, respectively, reflecting the human interactions with the grip
 of scissors and avoidance of
 the edge in teacher images.
 The inference model can therefore infer the interaction regions of
 human hands and scissors.

\begin{figure}
  {\centering
  \includegraphics[width=.35\textwidth]{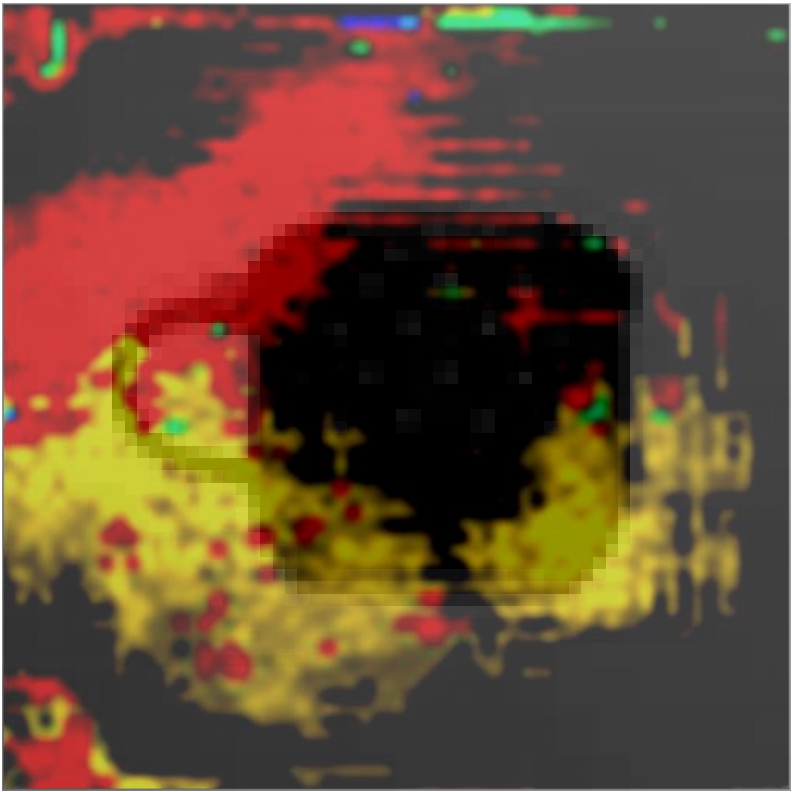}
  \caption{Inferred descriptors for interactions with handle (red) and
  bottom (yellow) of a cup}
  \label{fig:inferred_descriptors_for_each_part_of_a_cup}
  }
 \end{figure}

\begin{figure}
  {\centering
  \includegraphics[width=.35\textwidth]{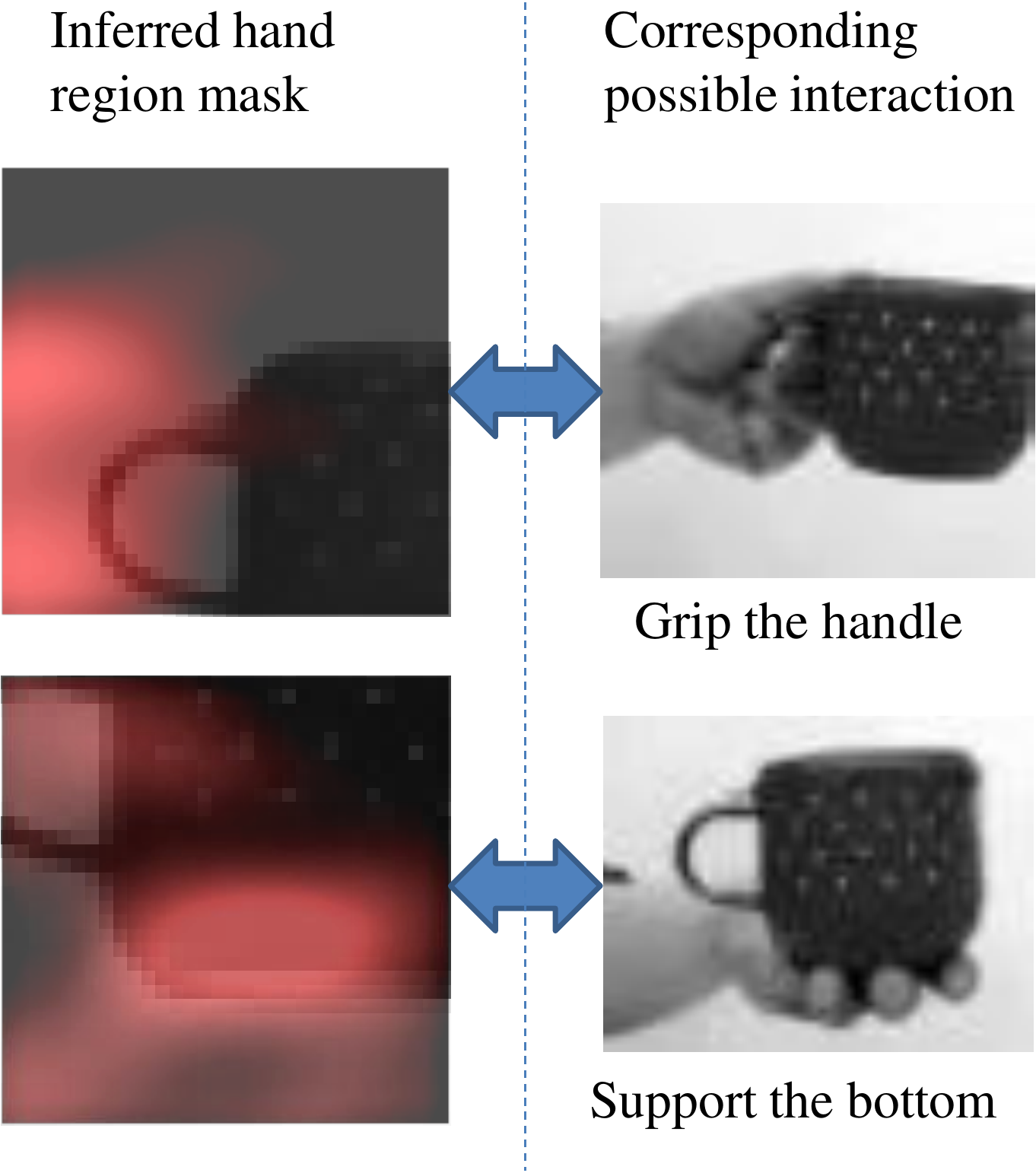}
  \caption{Inferred hand-region masks and their possible interaction}
  \label{fig:inferred_hand_region_mask_and_possible_interaction}
  }
 \end{figure}

 Figure \ref{fig:inferred_descriptors_for_each_part_of_a_cup} is a color
 representation of the
 inferred interaction descriptors for parts of a cup.
 The color of each position is determined by mean shift clustering of
 the vectors containing the inferred descriptor and the position.
 This figure reveals the different types of interactions inferred on
 the grip and the bottom of the cup.
 Figure ~\ref{fig:inferred_hand_region_mask_and_possible_interaction}
 shows the hand-region masks inferred in these two interaction types.
 The model can infer an interaction descriptor corresponding to possible
 interaction at a particular position.

\begin{figure}[t]
  {\centering
  \includegraphics[width=.35\textwidth]{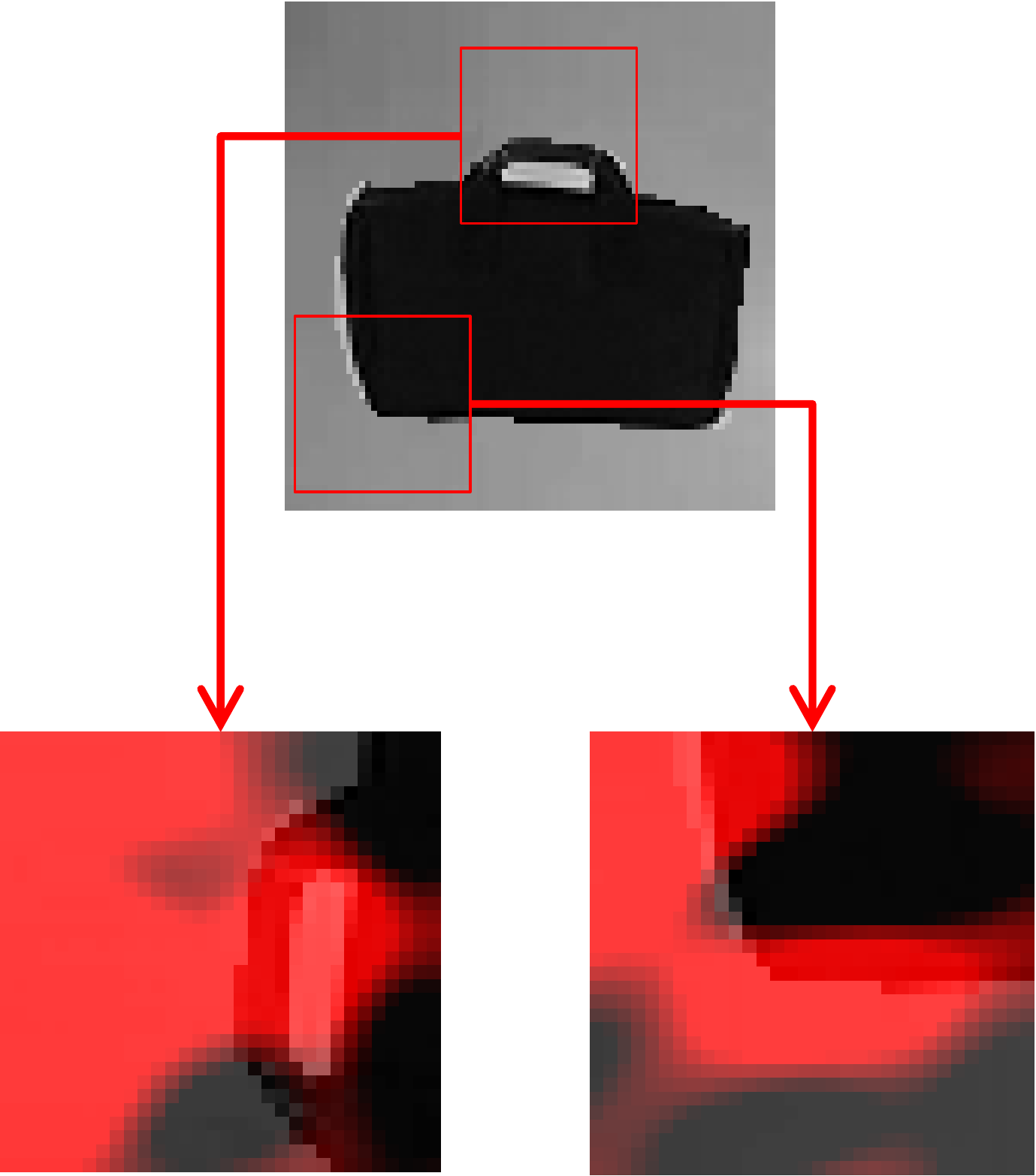}
  \revisedregion
  \caption{Hand-region masks inferred from an object in unknown category}
  \label{fig:inferred_hand_region_mask_and_possible_interaction2}
  }
 \end{figure}

 {\revisedregion
 To demonstrate that the proposed method can infer a possible
 interaction from an appearance of an unknown object in an unknown
 category,
 we inferred interaction descriptors from an appearance of a bag shown
 in the top image in
 Fig.~\ref{fig:inferred_hand_region_mask_and_possible_interaction2}.
 No bags are not used in training, but a grip of the bag has an
 appearance similar to that of a mug used in training.
 The bottom left image in
 Fig.~\ref{fig:inferred_hand_region_mask_and_possible_interaction2}
 is a hand region mask inferred from an image around a grip of the
 bag.
 It shows that the inference model can infer a hand region mask like
 handling the grip of the bag from an image rotated so that the
 direction of the grip is close to that of a mug used in training the
 inference model.
 This means that the inference model learns the relation between
 a grip-like shape and an interaction for handling it.
 And also, as shown in the bottom right image
 in Fig.~\ref{fig:inferred_hand_region_mask_and_possible_interaction2},
 a hand region like supporting the bottom is inferred from
 the other part of the bag, which is similar a bottom of a mug.

 This indicates that the inference model can infer a possible
 interaction from a partial appearance of an unknown object in an
 unknown category if the model is trained with similar partial
 appearances included in other objects.
 }

 {\revisedregion
 To evaluate the proposed inference model,
 we compared an interaction image inferred from an appearance of an
 object with a real instance of an interaction image occurred with
 the same object.
 To compare the two interaction images quantitatively, we calculated
 mean peak signal to noise ratios (PSNRs) between them for each
 channel (total appearance, hand region mask and object region mask),
 which is defined as below.
 \begin{equation}
  \frac{1}{N}
   \sum_{\left(I,I_{\textrm{obj}}\right)}
   10\log_{10}
   \frac
   {V_{c}^{2}}
   {\frac{1}{M}
   \norm{[I]_{c}
   - \left[D\left(R\left(I_{\textrm{obj}}\right)\right)\right]_{c}}_{\mathrm{L}2}^{2}},
 \end{equation}
 where
 \begin{equation}
  \begin{aligned}
   \left[I\right]_{c}
   &=
   (\textrm{the } c \textrm{-th channel of the interaction image } I),
   \\
   I_{\textrm{obj}}
   &= (\textrm{an object appearance}),
   \\
   I
   &= (\textrm{the interaction image  to } I_{\textrm{obj}}).
   \\
   N
   &= (\textrm{the number of samples}),
   \\
   M
   &= (\textrm{the number of pixels in a channel}),
   \\
   V_{c}
   &= (\textrm{the possible maximum pixel value}
   \\
   & \textrm{ of the }c\textrm{-th channel}).
  \end{aligned}
 \end{equation}
 The calculated values of PSNRs are shown in
 Tab.~\ref{tab:sn_ratio_of_inference}.
 The values of PSNRs for training samples are from 8[dB] to 10[dB].
 They are lower than 20[dB] that indicates unacceptable image
 quality in image compression\cite{1556624}.
 This is because
 the autoencoder extracts essential components common to some
 appearances of interactions instead of encoding detail of each
 interaction image.
 However,
 Fig.~\ref{fig:inference_result} shows that the proposed method can
 infer rough shapes of a hand and an object.
 From an appearance of a cutter, a hand mask region like grasping
 the cutter is inferred.
 An object hand mask region inferred from the cutter indicates
 a narrow and long region and it matches rough shape of the grip of
 the cutter.
 From an appearance of a cup, a hand mask region like grasping the
 cup is inferred.
 Although the value of PSNRs are not high, the proposed method can
 roughly infer a possible interaction.
 }

 \begin{table}[t]
  \begin{center}
   \revisedregion
   \caption{Mean PSNR of the inference model}
   \label{tab:sn_ratio_of_inference}
   \begin{tabular}{|c||c|c|c|}
    \hline
    & \multicolumn{3}{c|}{Mean PSNR [dB] for each channel}\\
    \cline{2-4}
    & Total appearance& Hand region& Object region\\
    \hline
    \begin{minipage}[c]{5em}
     for training\\
     samples
    \end{minipage}
    & 8.80& 10.33& 10.94\\
    \hline
    \begin{minipage}[c]{5em}
     for test\\
     samples
    \end{minipage}
    & 3.66& 6.53& 7.22\\
    \hline
   \end{tabular}
  \end{center}
 \end{table}

 \section{Conclusion}
 We proposed the {\em interaction descriptor space} for describing the
 hand--object interactions of functional objects.
 The space is automatically constructed from sets of object-handling
 images
 of typical functional objects such as mugs, scissors, cutters.
 We demonstrated that a descriptor corresponds to a quantitative
 interaction state and descriptors make clusters consistent with
 interaction types.
 We also proposed an inference model that infers a possible interaction
 from an object image alone.
 Given an object image, the model successfully inferred an interaction
 descriptor
 corresponding to a possible interaction at each position
 of the image.

 The interaction descriptor space can characterize hand--object
 interactions and it can be used to model the relations between an
 object and its
 possible interactions.
 The proposed approach is a potentially valuable tool in
 function-based classification.

\bibliographystyle{ieicetr}
\bibliography{mybibliography}


\profile[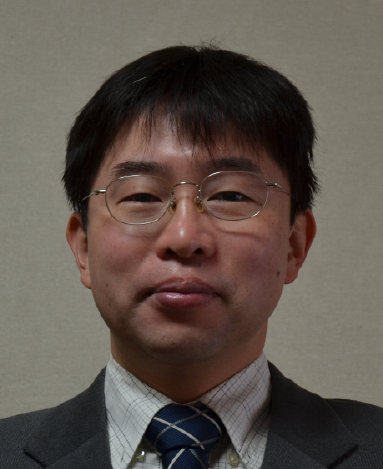]{Tadashi Matsuo}{
received the B.E., M.E. and Ph.D. degrees from Kyoto Institute of Technology
in 2001, 2003 and 2006, respectively.
He was a researcher in Research Organization of Science and Engineering,
Ritsumeikan University from 2006 to 2011.
He was an assistant in
College of Information Science and Engineering, Ritsumeikan University
in 2012, and is currently a TOKUNIN assistant professor.
His research interest includes image recognition and machine learning.
He is a member of IEICE.
}
\profile[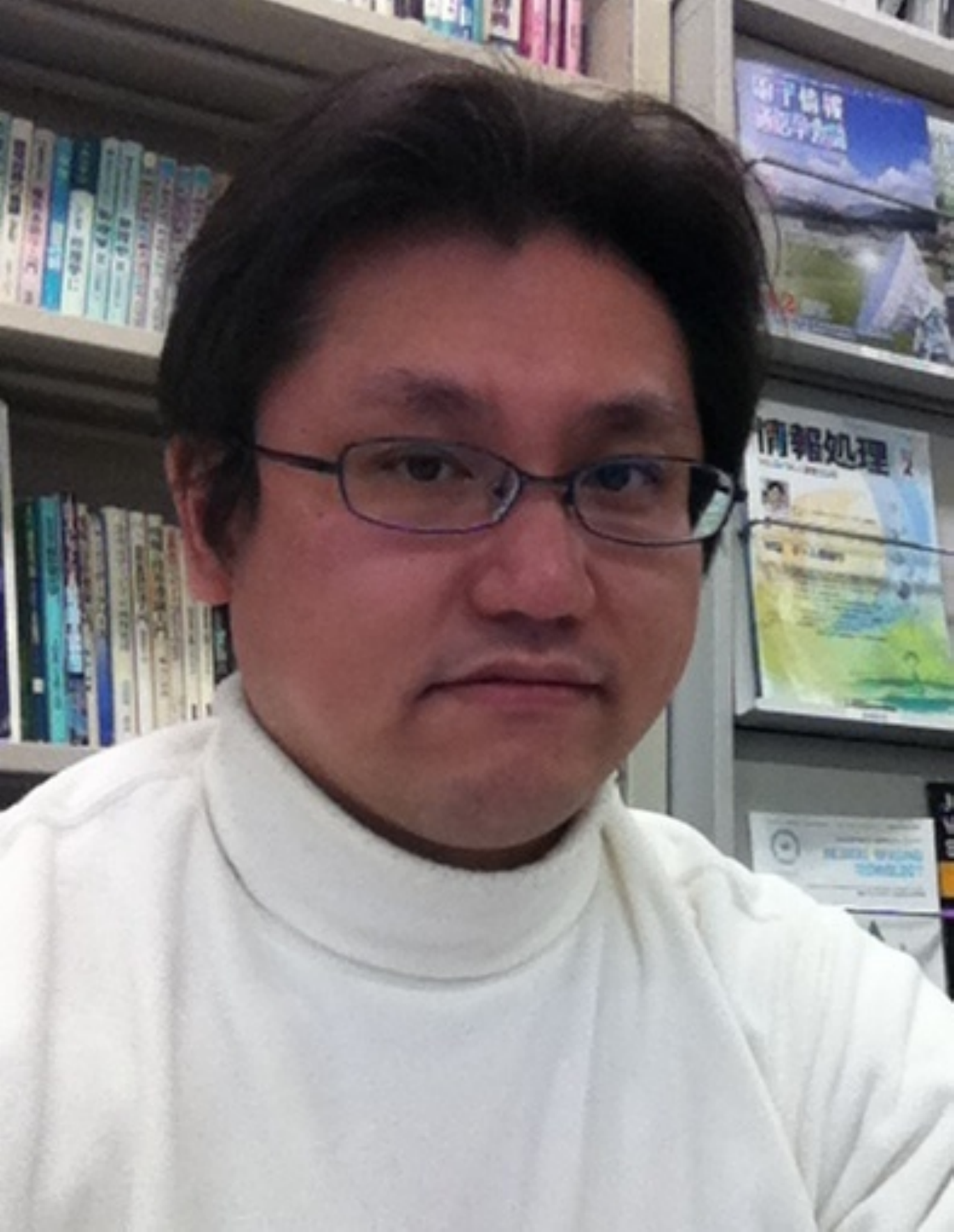]{Nobutaka Shimada}{
received the B.E., M.E. and Ph.D. degrees from Osaka University
in 1992, 1994 and 1997, respectively.
He was an assistant in Graduate School of Engineering, Osaka University
from 1997 to 2002.
In 2003, He was an associate professor in Graduate School of Engineering,
Osaka University.
He was an associate professor in
College of Information Science and Engineering, Ritsumeikan University
from 2004 to 2011, and is currently a professor.
In 2007, he was engaged in research as a visiting associate professor
in the Robotics Institute, Carnegie Mellon University.
His research interest includes computer vision, gesture interface and
interactive robots.
He is a member of IEEE, IEICE, IPSJ and JSAI.}

\end{document}